%% file: main.tex
\documentclass[10pt,twocolumn,letterpaper]{article}

\usepackage[pagenumbers]{cvpr} 

\input{preamble}
\usepackage{graphicx}
\usepackage{array}
\usepackage{booktabs}
\usepackage{multirow}
\usepackage{gensymb}
\usepackage[dvipsnames]{xcolor}
\usepackage{amsmath}
\definecolor{cvprblue}{rgb}{0.21,0.49,0.74}
\usepackage[pagebackref,breaklinks,colorlinks,allcolors=cvprblue]{hyperref}

\usepackage[capitalize]{cleveref}

\def\paperID{6377} 
\def\confName{CVPR}
\def\confYear{2025}

\title{{Neural LightRig}: Unlocking Accurate Object Normal and Material Estimation \\ with Multi-Light Diffusion}
\author{ Zexin~He\textsuperscript{1*}, 
 ~Tengfei~Wang\textsuperscript{2*}, 
 ~Xin~Huang\textsuperscript{2}, 
 ~Xingang~Pan\textsuperscript{3}, 
 ~Ziwei~Liu\textsuperscript{3} \\
$^1$The~Chinese~University~of~Hong~Kong, $^2$Shanghai~AI~Lab, $^3$Nanyang~Technological~University \\
}

\begin{document}

\twocolumn[{
\renewcommand\twocolumn[1][]{#1}%
\maketitle
\begin{center}
\vspace{-20pt}
    \centering
\includegraphics[width=0.99\linewidth]{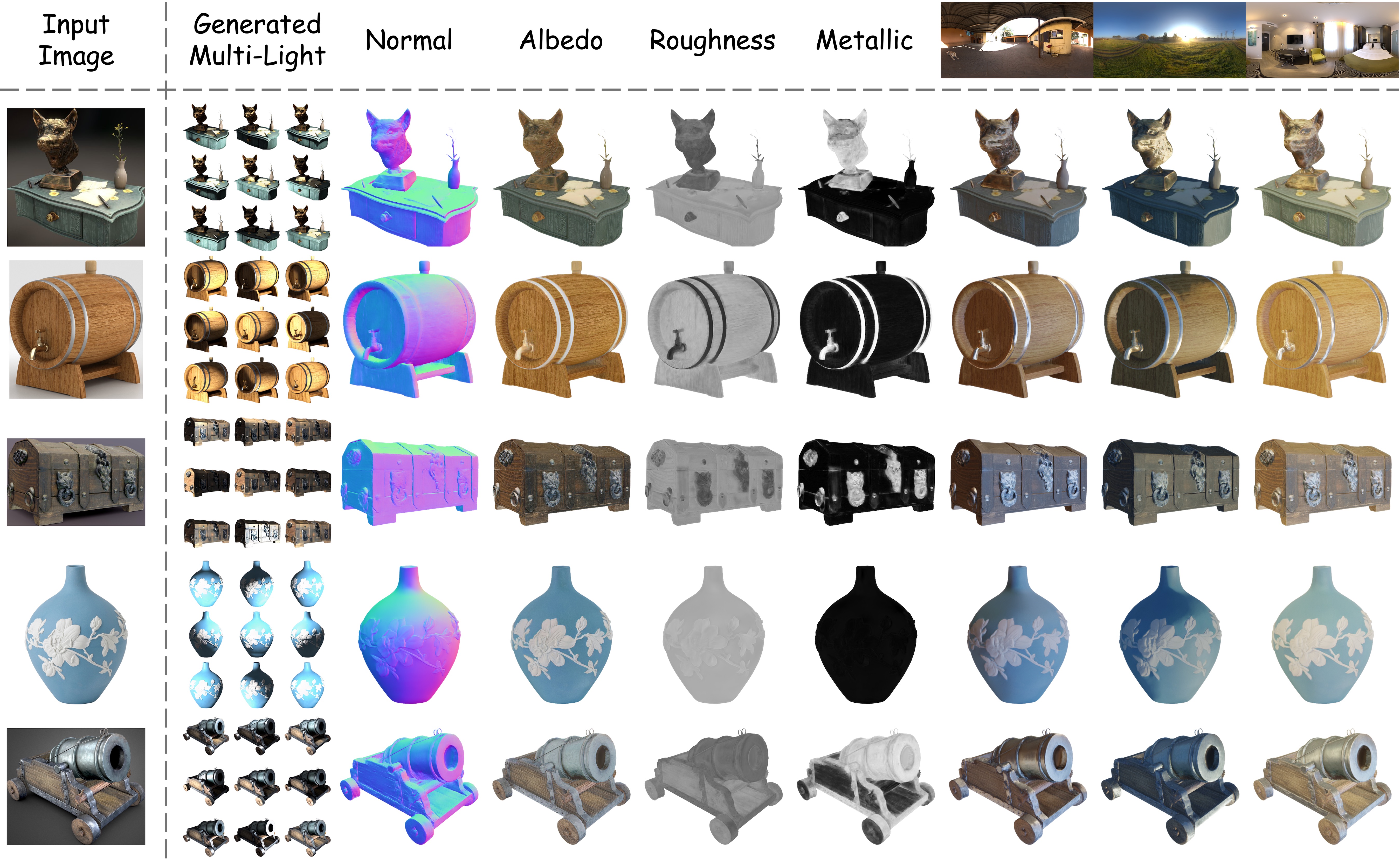}
\captionof{figure}{\emph{Neural LightRig} takes an image as input and generates multi-light images to assist the estimation of high-quality normal and PBR materials, which can be used to render realistic relit images under various environment lighting.}
\label{fig:teaser}
\end{center}}]

\let\thefootnote\relax\footnotetext{* Equal contribution. Work done during Zexin He's internship at Shanghai AI Lab.}

\input{Sections/0_abstract}
\input{Sections/1_intro}
\input{Sections/2_related}

\input{Sections/3_approach}
\input{Sections/4_exp}
\input{Sections/6_conclusion}

{
    \small
    \bibliographystyle{ieeenat_fullname}
    \bibliography{main}
}

\clearpage
\input{Sections/7_supp}

\end{document}

%% file: preamble.tex
\usepackage{cuted}

%% file: Sections/0_abstract.tex
\begin{abstract}
Recovering the geometry and materials of objects from a single image is challenging due to its under-constrained nature.  In this paper, we present \textbf{Neural LightRig}, a novel framework that boosts intrinsic estimation by leveraging auxiliary multi-lighting conditions from 2D diffusion priors. Specifically, \textbf{1)} we first leverage illumination priors from large-scale diffusion models to build our {multi-light diffusion model} on a synthetic relighting dataset with dedicated designs.
This diffusion model generates multiple consistent images, each illuminated by point light sources in different directions.  
\textbf{2)} By using these varied lighting images to reduce estimation uncertainty, we train a {large G-buffer model} with a U-Net backbone to accurately predict surface normals and materials.
Extensive experiments validate that our approach significantly outperforms state-of-the-art methods, enabling accurate surface normal and PBR material estimation with vivid relighting effects. 
Code and dataset are available on our project page at \href{https://projects.zxhezexin.com/neural-lightrig}{https://projects.zxhezexin.com/neural-lightrig}.
\end{abstract}
\vspace{-2mm}

%% file: Sections/1_intro.tex
\section{Introduction}\label{sec:introduction}

Recovering the geometry and physically-based rendering (PBR)  materials of real-world objects from images is a pivotal problem in graphics and computer vision. This task, also known as inverse rendering, facilitates a wide range of applications, such as video gaming, augmented and virtual reality, and robotics. In this paper, we proposed a data-driven approach for jointly estimating the surface normal and PBR materials of objects from a single image. Due to the complex interaction among geometry, materials, and environmental lighting, this ill-posed problem remains particularly challenging.

Prior research~\cite{barron2012shape,hasselgren2024nvdiffrecmc} has predominantly focused on optimization-based generation through differentiable rendering, which compares forward-rendered images with input images to refine normals and PBR materials. However, these methods are often time-consuming and heavily reliant on the capabilities of the differentiable renderer~\cite{Laine2020diffrast}. Though some works explored feed-forward estimation~\cite{zeng2024rgbx,liu2020unsupervised,yi2023weaklysupervised}, their quality and generalizability still remain challenging, due to the inherently ill-posed nature of inferring geometry and materials from a single image.

For precise normal and material acquisition, photometric stereo techniques~\cite{woodham1989photostereo} are widely employed, as they mitigate ambiguity by capturing multiple images from the same viewpoint with various lighting. 
These images are illuminated by different point light sources,  
which provide variations in surface reflectance to enrich information. However, such methods~\cite{drbohlav2005calibrateps,debevec2000acquiring,levoy2000michelangelo} often require complex capture systems with sophisticated cameras or lighting setups, which can be costly and impractical for in-the-wild images. Given the promising advances in image diffusion models, we ask the question: can we develop a multi-light diffusion model to simulate images illuminated by different directional light sources, thereby improving surface normal and material estimation (as shown in \cref{fig:teaser})? 

Our motivation arises from recent advances in 3D generation, which employ diffusion models~\cite{shi2023zero123plus,li2024instant3d} to generate multi-view images and train reconstruction models~\cite{hong2024lrm} for 3D reconstruction. These multi-view diffusion models have demonstrated the potential to manipulate camera views of pre-trained image diffusion models 
such as Stable Diffusion~\cite{rombach2022ldm}. Similarly, we aim to expand the use of pre-trained diffusion models for multi-light image generation.

In this work, we present \emph{Neural LightRig} for joint normal and material estimation of objects from monocular images, which consists of a multi-light diffusion model and a large prediction model. Given an input image, the \textbf{multi-light diffusion model} produces consistent and high-quality relit images under various point light sources (as shown in \cref{fig:method_lighting_relation}). To achieve this, we create a synthetic relighting dataset for training with Blender~\cite{blender}. With a dedicated architecture and training design, our diffusion model enables the multi-light generation of objects from arbitrary categories. The \textbf{ large G-buffer model} then processes the generated multi-light images to produce surface normals and PBR materials, such as albedo, roughness, and metallic. We employ a UNet architecture for efficient and high-resolution prediction, with end-to-end supervision at the pixel level. To bridge the domain gap between multi-light images rendered from 3D objects and those generated by diffusion models, we further design a series of data augmentation strategies for domain alignment.

\begin{figure*}[t]
    \centering
    \includegraphics[width=\linewidth]{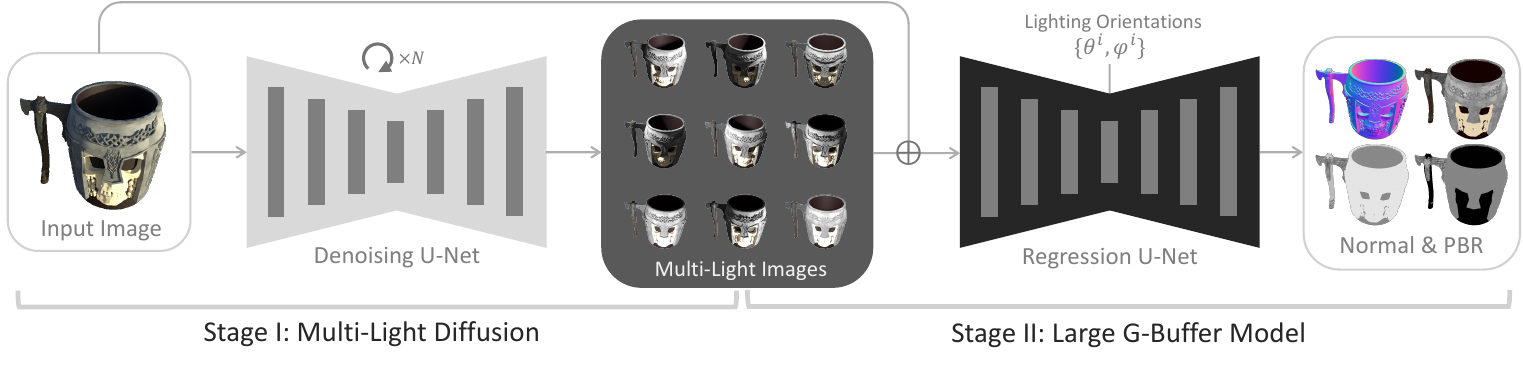}
    \caption{\textbf{Framework Overview.} Multi-light diffusion generates multi-light images from an input image. These images with corresponding lighting orientations are then used to predict surface normals and PBR materials with a regression U-Net.}
    \label{fig:method_overview}
\end{figure*}

Taken together, the proposed framework demonstrates remarkable performance on both synthetic and real-world images. Extensive qualitative and quantitative evaluations show that \emph{Neural LightRig} surpasses existing approaches in surface normal estimation, PBR material estimation, and single-image relighting.
Comprehensive visual results are provided in the appendix and on our \href{https://projects.zxhezexin.com/neural-lightrig}{project page}.
Our key contributions are as follows:
\begin{itemize}
    \item We propose a novel approach for object normal and PBR estimation from monocular images, reformulating this ill-posed problem by simulating multi-lighting conditions.
    \item We construct a synthetic dataset for multi-light image generation and surface property estimation. With this dataset, we demonstrate the capability to manipulate diffusion models for consistent multi-light generation.
    \item Extensive experiments validate the effectiveness of our method, establishing new state-of-the-art results.
\end{itemize}

%% file: Sections/2_related.tex
\section{Related Works}
\noindent\textbf{Diffusion Models.}
Well-trained diffusion models~\cite{rombach2022ldm,wang2022pretraining} have shown promising potential in providing essential priors for under-determined tasks.
Recent works showcase the utility of image diffusion models in novel-view  synthesis~\cite{liu2023zero123,shi2023zero123plus,wang2024phidias,liu2024syncdreamer,shi2024mvdream,liu2024one2345plus}, which combines with reconstruction models~\cite{hong2024lrm,openlrm,tang2024lgm} to achieve high-quality 3D generation. Similarly, some recent works attempt to leverage the learned priors in diffusion models to simulate lighting variations~\cite{jin2024neural_gaffer,zeng2024dilightnet}, but they do not account for the consistency of multi-light generation. In contrast, we aim to generate multiple images under different lighting sources that facilitate object surface property estimation.

\noindent\textbf{Monocular Normal Estimation.}
Estimating surface normals from a single image is a classic yet under-determined problem.
Early works often relied on photometric cues or handcrafted features~\cite{hoiem2005automatic,hoiem2007recovering,fouhey2013primitives}, while later works adopted deep learning to improve accuracy~\cite{ladicky2014discriminatively,li2015regressioncrf,bansal2016marr,zhang2019patternaffi,do2020tilted,wang2020vplnet,qi2022geonetplus,xu2024genpercept}. More recently, large-scale datasets~\cite{objaverse, eftekhar2021omnidata} have further advanced regression-based methods~\cite{bae2021uncertainty,baradad2023depthprompt,bae2024dsine}. 
Despite promising results, they struggle with complex details due to inherent ambiguity.
Diffusion-based methods~\cite{fu2024geowizard,ke2023marigold,ye2024stablenormal},  turn to generative priors~\cite{rombach2022ldm} to help address such ambiguity but often fall short in accurately aligning with ground truth, leading to deviations in finer geometric details crucial for downstream tasks.

\noindent\textbf{Material Estimation.}
Material estimation aims to recover intrinsic properties from images, which is an ill-posed problem,
as multiple combinations of materials and lighting conditions could lead to the same appearance, 
Traditional methods attempted to employ photometric stereos~\cite{woodham1989photostereo,drbohlav2005calibrateps} to disambiguate this problem under controlled lighting conditions~\cite{debevec2000acquiring,levoy2000michelangelo}. Some works~\cite{boss2021nerd,srinivasan2021nerv,zhang2023neilfplus,hasselgren2024nvdiffrecmc} optimize neural representation with multi-view images.
Later, the emergence of large-scale synthetic datasets~\cite{vecchio2023matsynth,objaverse} has advanced data-driven approaches~\cite{shi2017nonlambertian,lichy2021capturehome,sang2020singleshot,yu2019inverserendernet,yi2023weaklysupervised,liu2020unsupervised}, but they still contend with under-determination. Recently, diffusion-based methods~\cite{lyu2023dpi,zeng2024rgbx,chen2024intrinsicanything,huang2024materialanything} have emerged as a promising alternative, but often suffer from domain shift between material images and natural images.

%% file: Sections/3_approach.tex
\section{Approach}
Given an image $\mathbf{I}$, we aim to estimate both its surface normal $\mathbf{n}$ and PBR materials (albedo $\mathbf{a}$, roughness $\mathbf{r}$, and metallic $\mathbf{m}$), where $\mathbf{n},\mathbf{a} \in \mathbb{R}^{H\times W\times3}$ and $\mathbf{r},\mathbf{m} \in \mathbb{R}^{H\times W\times1}$.
These surface properties, commonly known as G-buffers in graphics, are collectively denoted as $\mathcal{B}=\{\mathbf{n,a,r,m}\}$. However, interpreting these properties from a single lighting condition is challenging due to the the under-constrained nature of the problem. To address this, we propose \emph{Neural LightRig}, as illustrated in \cref{fig:method_overview}. Our approach leverages a multi-light diffusion model (\cref{sec:multi_light_diffusion}) to generate multi-light images from the input, which then act as enriched conditions to alleviate the inherent ambiguity in G-buffer prediction model (\cref{sec:large_gbuffer_model}). We further describe the construction of our synthetic dataset, \emph{LightProp}, which supports both stages of our framework, in \cref{sec:method_dataset}.

\subsection{Multi-Light Diffusion}
\label{sec:multi_light_diffusion}
To obtain surface reflectance variations that increase contextual information for accurate G-buffer estimation, we learn a diffusion model $g(\cdot)$ to generate $L$ multi-light images from the input image $\mathbf{I}$:
\begin{equation}
    \{\mathbf{x}^i \mid i = 1, 2, \dots, L\} = g(\mathbf{I}).
\end{equation}
In particular, we set $L=9$ to balance performance and efficiency, covering a diverse range of lighting variations (\cref{fig:method_lighting_relation}) without excessive overhead.

\noindent\textbf{Generating Multi-Light Images.}
Collecting such training pairs is challenging due to the limited availability of 3D objects with PBR~\cite{vecchio2023matsynth,objaverse} and the high cost of real-world capturing in photometric stereos~\cite{kaya2023mvpsrevisited}. 
Fortunately, diffusion models trained on massive internet images have shown an inherent ability to model complex 3D shapes and textures, which have been applied for novel view synthesis~\cite{shi2023zero123plus} and relighting~\cite{zeng2024dilightnet,jin2024neural_gaffer}.
We thus leverage the prior from a well-trained image diffusion model and fine-tune it for multi-light generation, arguing that such a well-trained image generation model possesses the capacity to simulate diverse lighting conditions. 
Rather than generating each-light image $\mathbf{x}^i$ separately, we arrange nine-light images in a $3\times3$ grid layout to form a single image $\mathbf{x}$, allowing the simultaneous generation for them. This simple configuration facilitates efficient cross-image context communication, thereby enhancing the consistency of generated multi-light images.

\begin{figure}[t]
    \centering
    \includegraphics[width=\linewidth]{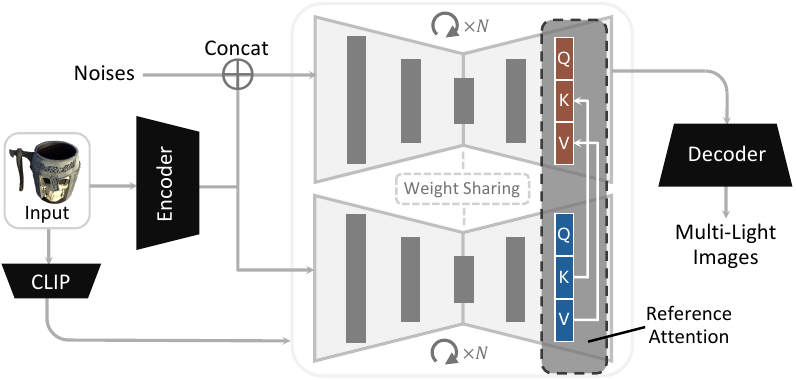}
    \caption{{Hybrid condition in multi-light diffusion}. Input images are incorporated via \textit{concatenation} with noise latents and enhanced through \textit{reference attention}, where queries in the denoise stream attend to keys and values from both streams.}
    \label{fig:method_conditioning}
\end{figure}

\noindent\textbf{Conditioning Strategy.}
To incorporate the input image into the diffusion model, we employ a hybrid conditioning method, as illustrated in \cref{fig:method_conditioning}.
As the input images are pixel-wise aligned with the multi-light images, we naturally apply \textit{channel-wise concatenation}.
This straightforward concatenation effectively captures the variations between the input and each multi-light image, which is essential for generating accurate lighting effects.
However, we found this simple concatenation alone is inadequate for generating high-fidelity multi-light images, leading to discrepancies in color tone and texture relative to the input.
To address this, we further adopt \textit{reference attention}~\cite{refattn,shi2023zero123plus}, where self-attention layers in the denoising U-Net also attend to keys and values obtained from the input image. This is represented as $\mathrm{Attn}(\textcolor{brown}{\mathbf{Q}}, [\textcolor{brown}{\mathbf{K}},\textcolor{NavyBlue}{\mathbf{K}_\text{cond}}], [\textcolor{brown}{\mathbf{V}},\textcolor{NavyBlue}{\mathbf{V}_\text{cond}}])$,
in which $\mathbf{Q,K,V}$ are the query, key, and value tokens from the denoising stream, and the subscript $\text{``cond"}$ denotes tokens from the input image.
This combined approach manages to preserve desired textures from in the input and is crucial for generating high-quality and realistic multi-light images.

\noindent\textbf{Tuning Scheme.}
We build our model on Stable Diffusion \emph{v}-version model~\cite{sd2_1,salimans2022vprediction}. Let $\alpha_t, \sigma_t$ be the controlling factors in the diffusion process, and define ground-truth velocity as $\mathbf{v} = \alpha_t \mathbf{\epsilon} + \sigma_t \mathbf{x}$ and predicted velocity as $\mathbf{v}_{\theta}(\cdot)$. The training target can be denoted as:
\begin{equation}
    \mathcal{L} = \mathbb{E}_{\mathbf{x}, \mathbf{I}, \mathbf{\epsilon}, t} \left[\| \mathbf{v} - \mathbf{v}_{\theta}(\mathbf{z}_t, t, \mathbf{I}) \|^2 \right],
\end{equation}
where $\mathbf{z}_t$ is the noisy latent of $\mathbf{x}$ at timestep $t$, and $\mathbf{I}$ is the input image.
To fully leverage the capacity of diffusion model, we adopt a two-phase training scheme. Initially, we freeze most parameters except for the first convolution layer and all attention layers to warm up the weights. This stabilizes early training, allowing for a smooth transition without severely disrupting the pre-trained model. Afterwards, we fine-tune the entire model at a considerably lower learning rate, facilitating careful adaptation for multi-light generation while retaining as much prior knowledge as possible.

\subsection{Large G-Buffer Model}
\label{sec:large_gbuffer_model}
Next, we learn a regression model $f(\cdot)$ to predict normals and PBR maps with the auxiliary multi-light images. 

\begin{figure}[t]
    \centering
    \includegraphics[width=0.9\linewidth]{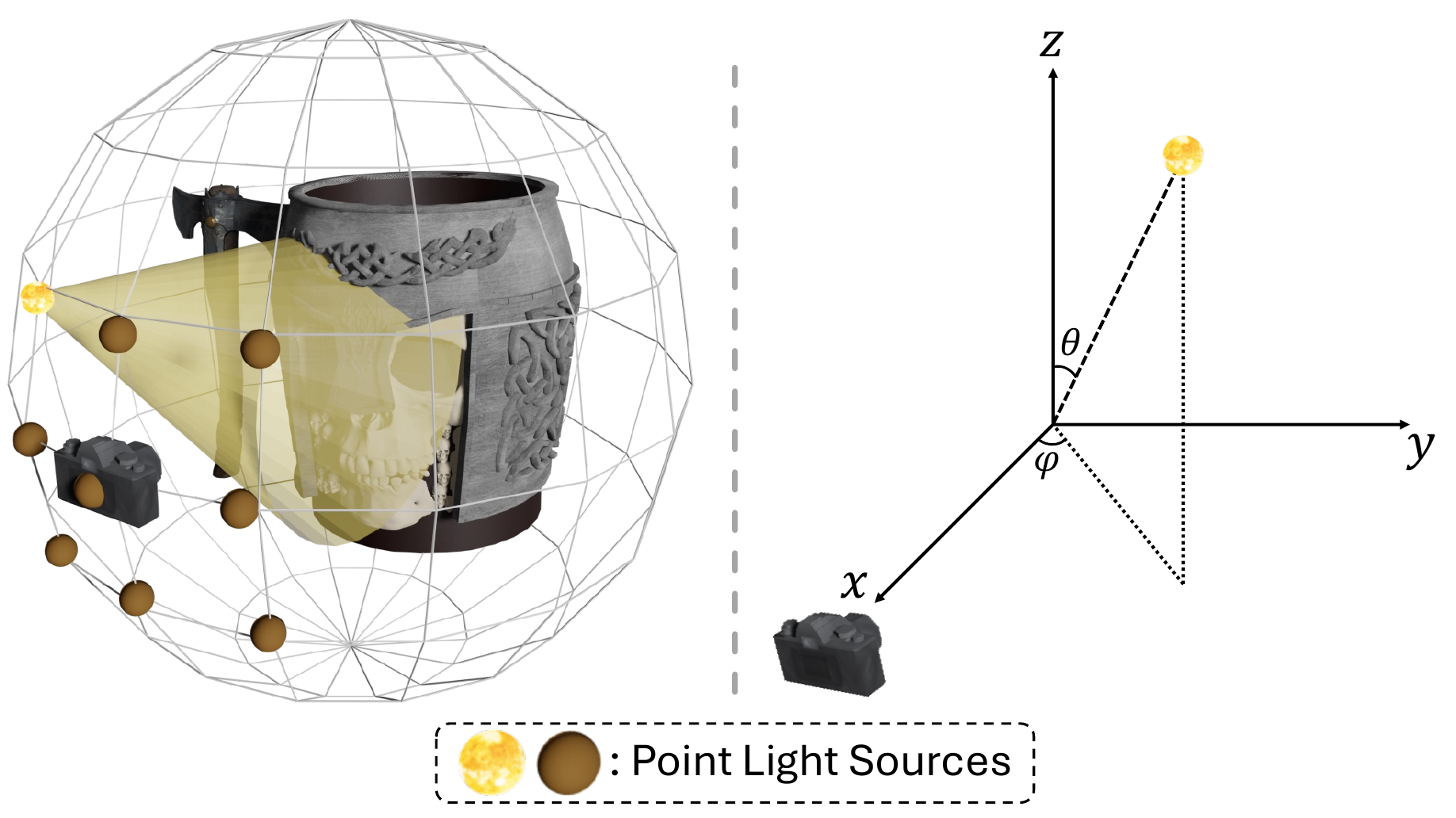}
    \caption{Visualization of multi-light setup in \textit{LightProp}. Camera and point lights are positioned on a sphere around the object. $\theta, \varphi$ are spherical coordinates to determine each light's orientation relative to the object.}
    \label{fig:method_lighting_relation}
\end{figure}

\noindent\textbf{Prediction Model.} 
Since the input image, multi-light images, and G-buffer maps are pixel-wise aligned, we opt for a U-Net architecture thanks to its efficiency in high-resolution prediction. Also, U-Net provides inductive bias for learning spatial relations, making it well-suited for our task. The model takes channel-wise concatenated input and multi-light images, and outputs an $8$-channel G-buffer, containing $3$-channel $\mathbf{n}$ and $\mathbf{a}$ maps, and $1$-channel $\mathbf{r}$ and $\mathbf{m}$ maps.
This multi-light-enhanced G-buffer prediction is represented as:
\begin{equation}
    \mathcal{B} = f\left(\mathbf{I}, \left\{(\mathbf{x}^i, \theta^i, \varphi^i) \mid i = 1, 2, \dots, L\right\}\right),
\end{equation}
where each novel-light image $\mathbf{x}^i$ is associated with the light source poses  $\theta^i$ and $\varphi^i$, which indicate spherical coordinates of the light source relative to the object (see \cref{fig:method_lighting_relation}).
Conditioning on these poses allows $f(\cdot)$ to explicitly correlate shading variations with their respective light sources, enhancing surface estimation.

\noindent\textbf{Training Objectives.} 
To train the model $f(\cdot)$ for G-buffer prediction, we apply loss functions to each of the G-buffer properties. 
We employ a cosine similarity loss for normals, enforcing the model to capture precise surface orientations. 
To stabilize the training, we also include an MSE term as regularization:
\begin{equation}
    \mathcal{L}_{\text{normal}} = \left( 1 - \frac{\mathbf{n} \cdot \hat{\mathbf{n}}}{\|\mathbf{n}\| \|\hat{\mathbf{n}}\|} \right) + \lambda_{1}\| \mathbf{n} - \hat{\mathbf{n}} \| ^2 ,
\end{equation}
where $\hat{\mathbf{n}}$ and $\mathbf{n}$ are the predicted and ground-truth normals. 
For the predicted albedo $\hat{\mathbf{a}}$, roughness $\hat{\mathbf{r}}$, and metallic $\hat{\mathbf{m}}$, we simply use MSE losses as:
\begin{equation}
    \mathcal{L_{\text{PBR}}} = \| \mathbf{a} - \hat{\mathbf{a}} \|^2 + \| \mathbf{r} - \hat{\mathbf{r}} \|^2 + \| \mathbf{m} - \hat{\mathbf{m}} \|^2.
\end{equation}
The overall loss is the weighted sum of the two losses.

\noindent\textbf{Augmentations.}
We train our prediction model using ground-truth rendered multi-light images, but for inference, we rely on generated images from diffusion models. In our earlier experiments, we observed a domain gap between the generated and rendered multi-light images in  sharpness and brightness. This gap would introduce discrepancies between training and inference, causing degraded performances. To bridge this gap, we apply a series of augmentations to multi-light images during training, including:
(a) \textit{Random Degradation}, such as resizing and grid distortion that simulate small misalignments;
(b) \textit{Random Intensity} that adjusts brightness in HSV space, simulating brightness variations of multi-light images;
(c) \textit{Random Orientation} perturbs $\{\theta^i, \varphi^i\}$ to account for potential disparities, encouraging $f(\cdot)$ to be robust to inaccurate lighting cues; and
(d) \textit{Data Mixing}, where we mix generated multi-light images into the training data to further mitigate this gap.

\begin{table*}[t]
\centering
  \caption{Quantitative comparison on surface normal estimation. We report mean and median angular errors, as well as accuracies within different angular thresholds from $3\degree$ to $30\degree$.}
  \label{tab:comparison_full_normal}
  \scriptsize
  \begin{tabular}{l@{\hspace{9mm}} |@{\hspace{7mm}} c@{\hspace{7mm}}c@{\hspace{7mm}} |@{\hspace{7mm}} c@{\hspace{10mm}}c@{\hspace{10mm}}c@{\hspace{10mm}}c@{\hspace{10mm}}c@{\hspace{10mm}}c}
    \toprule
    \textbf{Method}& \textbf{Mean~$\downarrow$} & \textbf{Median~$\downarrow$} & \textbf{3\degree~$\uparrow$} & \textbf{5\degree~$\uparrow$} & \textbf{7.5\degree~$\uparrow$} & \textbf{11.25\degree~$\uparrow$} & \textbf{22.5\degree~$\uparrow$} & \textbf{30\degree~$\uparrow$}
    \\
    \midrule
    RGB$\leftrightarrow$X~\cite{zeng2024rgbx} & 14.847\hspace{1.3mm} & 13.704\hspace{1.3mm} & 11.676 & 23.073 & 35.196 & 49.829 & 75.777 & 86.348
    \\
    DSINE~\cite{bae2024dsine}  & 9.161 & 7.457 & 23.565 & 41.751 & 57.596 & 72.003 & 90.294 & 95.297
    \\
    GeoWizard~\cite{fu2024geowizard}  & 8.455 & 6.926 & 22.245 & 40.993 & 58.457 & 74.916 & 93.315 & \underline{97.162}
    \\
    Marigold~\cite{ke2023marigold}  & 8.652 & 7.078 & \underline{25.219} & 42.289 & 58.062 & 72.873 & 92.326 & 96.742
    \\
    StableNormal~\cite{ye2024stablenormal}  & \underline{8.034} & \underline{6.568} & 21.393 & \underline{43.917} & \underline{63.740} & \underline{78.568} & \underline{93.671} & 96.785
    \\
    \midrule
    \textbf{Ours}  & \textbf{6.413} & \textbf{4.897} & \textbf{38.656} & \textbf{56.780} & \textbf{70.938} & \textbf{82.853} & \textbf{95.412} & \textbf{98.063}
    \\
  \bottomrule
\end{tabular}
\end{table*}

\begin{table*}[t]
\centering
  \caption{Quantitative comparison on PBR materials estimation and single-image relighting.}
  \label{tab:comparison_full_pbr_relit}
  \scriptsize
  \begin{tabular}{l@{\hspace{8mm}} |@{\hspace{3mm}} cc |cc  | cc | ccc | c}
    \toprule
    \textbf{Method} & \multicolumn{2}{c|}{\textbf{Albedo}} & \multicolumn{2}{c|}{\textbf{Roughness}} & \multicolumn{2}{c|}{\textbf{Metallic}} & \multicolumn{3}{c|}{\textbf{Relighting}} & \multicolumn{1}{c}{\textbf{Latency}}
    \\
    & \textbf{PSNR~$\uparrow$} & \textbf{RMSE~$\downarrow$} & \textbf{PSNR~$\uparrow$} & \textbf{RMSE~$\downarrow$} & \textbf{PSNR~$\uparrow$} & \textbf{RMSE~$\downarrow$} & \textbf{PSNR~$\uparrow$} & \textbf{SSIM~$\uparrow$} & \textbf{LPIPS~$\downarrow$} & \textbf{Average Time~$\downarrow$}
    \\
    \midrule
    RGB$\leftrightarrow$X~\cite{zeng2024rgbx} & 16.26 & 0.176 & \underline{19.21} & \underline{0.134} & 16.65 & 0.199 & 20.78 & 0.8927 & 0.0781 & 15s
    \\
    Yi. et al~\cite{yi2023weaklysupervised}  & 21.10 & 0.106 & 16.88 & 0.180 & 20.30 & 0.144 & 26.47 & 0.9316 & 0.0691 & 5s
    \\
    IntrinsicAnything~\cite{chen2024intrinsicanything}  & \underline{23.88} & \underline{0.078} & 17.25 & 0.172 & \underline{22.00} & \underline{0.134} & \underline{27.98} & \underline{0.9474} & \underline{0.0490} & 2min
    \\
    DiLightNet~\cite{zeng2024dilightnet}  & - & - & - & - & - & - & 22.68 & 0.8751 & 0.0981 & 30s
    \\
    IC-Light~\cite{iclight}  & - & - & - & - & - & - & 20.29 & 0.9027 & 0.0638 & 1min
    \\    
    \midrule
    \textbf{Ours}  & \textbf{26.62} & \textbf{0.054} & \textbf{23.44} & \textbf{0.085} & \textbf{26.23} & \textbf{0.109} & \textbf{30.12} & \textbf{0.9601} & \textbf{0.0371} & 5s
    \\
  \bottomrule
\end{tabular}
\vspace{-3mm}
\end{table*}

\subsection{LightProp Dataset}
\label{sec:method_dataset}
To train our model, we need to collect paired multi-light images and corresponding normal and PBR material maps. However, capturing such pairs in the real world requires specialized photometric equipments and controlled lighting, which is impractical for large-scale collection, while internet images typically lack access to their underlying 3D data, making it infeasible to derive ground-truth surface properties.
Therefore, we construct a synthetic dataset \textit{LightProp}, where we curate $80k$ objects from Objaverse~\cite{objaverse}, filtering out those of low-quality or without PBR materials.

\textit{LightProp} provides multi-light images and G-buffer maps for every object. Each object is rendered at $5$ random views, and for each view, we simulate $5$ images under random lighting conditions, including point light, area light, and HDR environment maps. Each view also provides a full set of surface normal and PBR materials, along with multi-light images rendered under known directional lighting. As shown in \cref{fig:method_lighting_relation}, we position the camera and point lights on a sphere around the object, where $\theta$ determines the vertical position of the lights relative to the overhead direction, and $\varphi$ controls the rotation relative to the camera. In practice, the positions of light sources are fixed during the training of multi-light diffusion model $g(\cdot)$ and the inference of G-buffer prediction model $f(\cdot)$, while randomized light positions are applied for training $f(\cdot)$ to encourage generalization. More details on dataset construction can be found in the \emph{appendix}.

%% file: Sections/4_exp.tex
\section{Experiments}

\begin{figure*}[t]
    \centering
    \includegraphics[width=\linewidth]{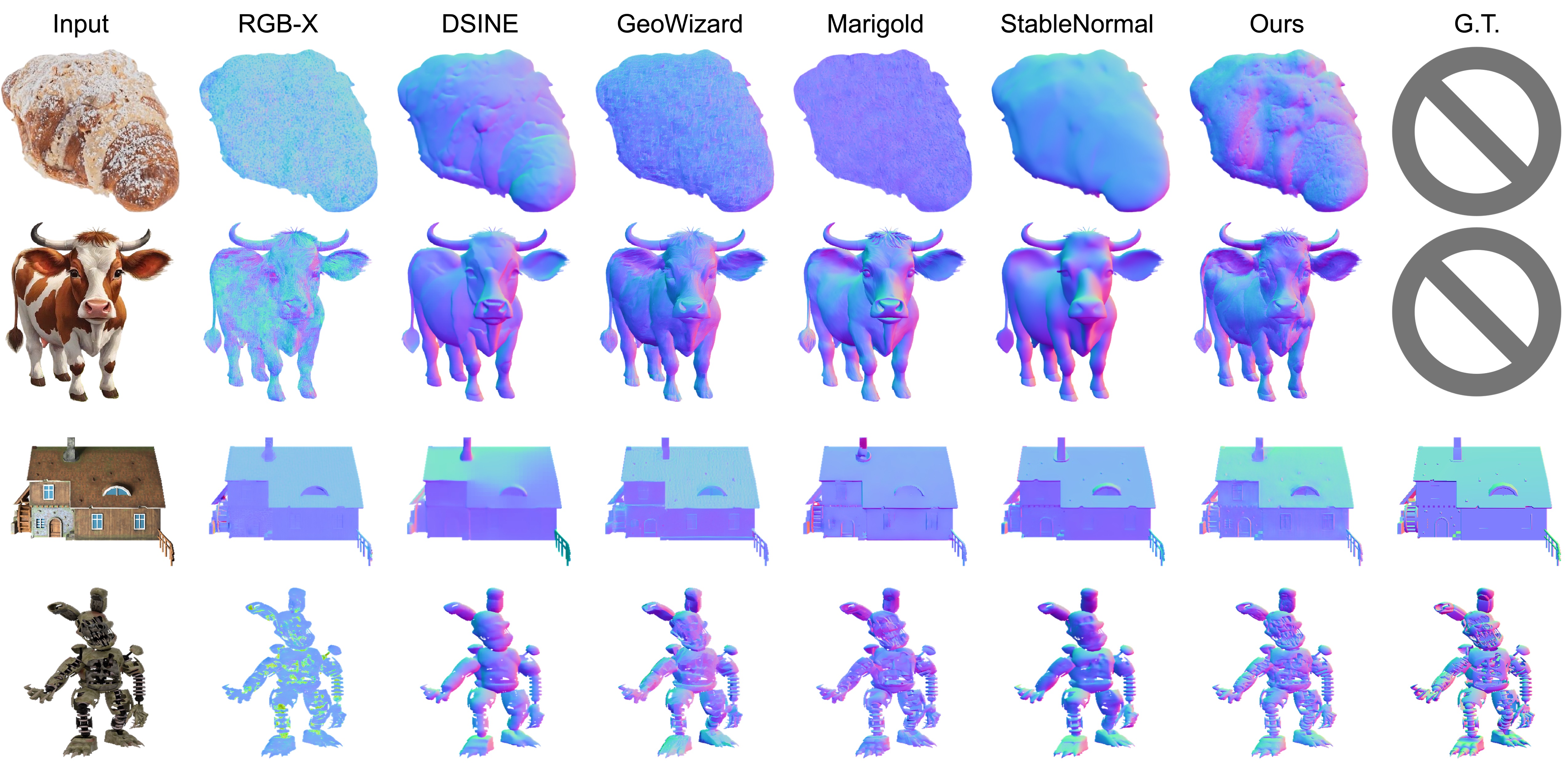}
    \caption{Qualitative comparison on surface normal estimation.  Ground truth normals (G.T.) are provided for input images rendered from available 3D objects (the last two rows) and are omitted for in-the-wild images (the first two rows).}
    \label{fig:normal_compare_main}
\end{figure*}

\begin{figure*}[t]
    \centering
    \includegraphics[width=\linewidth]{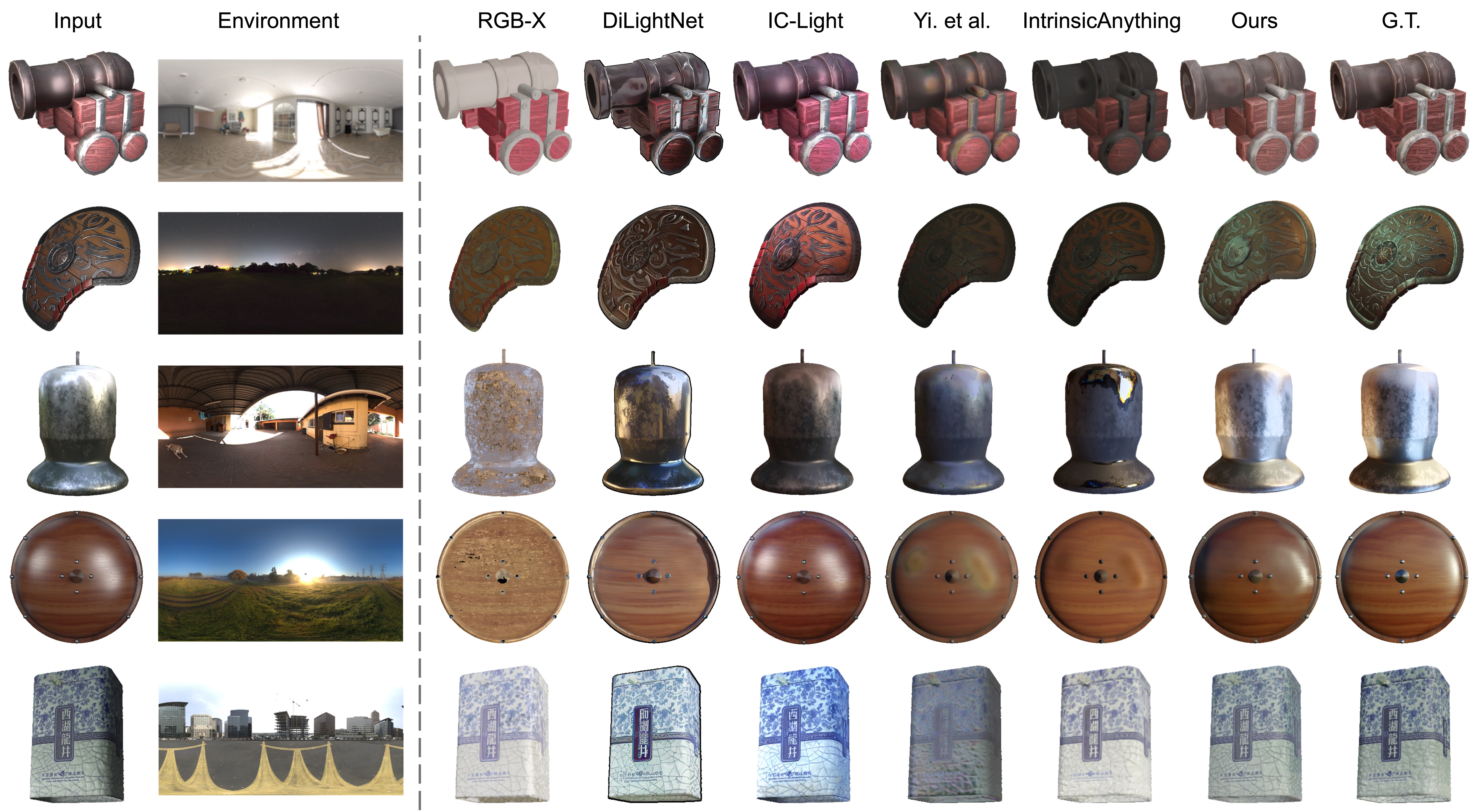}
    \caption{Qualitative comparison on single-image relighting.}
    \label{fig:relit_compare_main}
    \vspace{-2mm}
\end{figure*}

We evaluate our method across various tasks.
For \textbf{normal estimation}, we benchmark against regression-based method DSINE~\cite{bae2024dsine} and diffusion-based methods  GeoWizard~\cite{fu2024geowizard}, Marigold~\cite{ke2023marigold} and StableNormal~\cite{ye2024stablenormal}.
For \textbf{PBR material prediction},  we compare our method with a data-driven method by \citet{yi2023weaklysupervised}, an optimization method IntrinsicAnything~\citep{chen2024intrinsicanything}, and a diffusion-based model RGB$\leftrightarrow$X~\cite{zeng2024rgbx}.
For \textbf{image relighting}, we use ground-truth normal maps and predicted PBR materials from baselines~\cite{yi2023weaklysupervised,chen2024intrinsicanything,zeng2024rgbx} to render relit images, serving as relighting baselines. We also compare our method with diffusion-based image relighting models DiLightNet~\cite{zeng2024dilightnet} and IC-Light~\cite{iclight}, using a captioning model~\cite{blip2} to generate prompts.

\begin{figure*}[t]
    \centering
    \includegraphics[width=\linewidth]{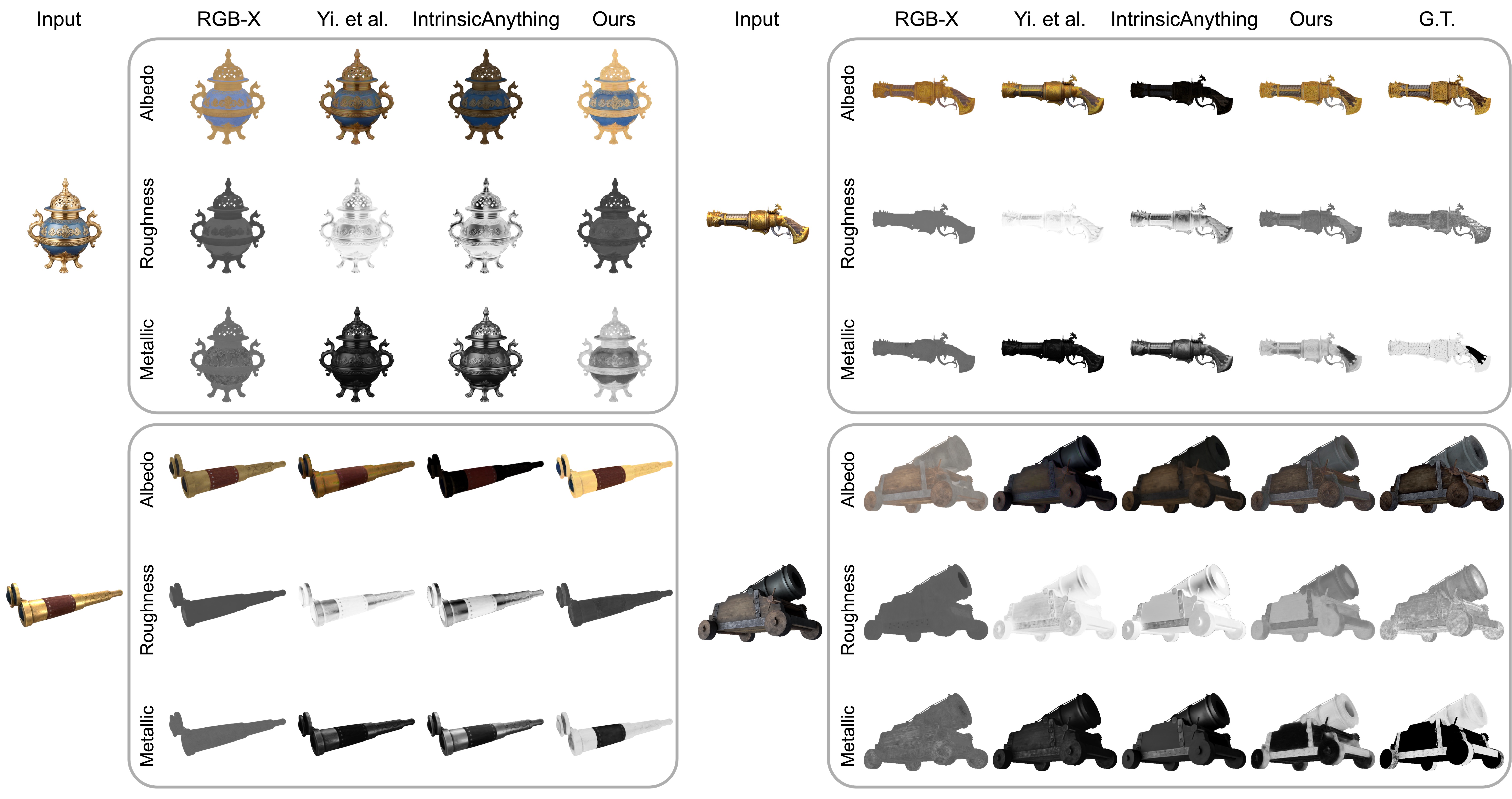}
    \caption{Qualitative comparison on PBR material estimation. Ground truth materials (G.T.) are provided for input images rendered from available 3D objects (the right column) and are omitted for in-the-wild images (the left column).}
    \label{fig:pbr_compare_main}
\end{figure*}

\subsection{Quantitative Evaluation}
We calculate metrics on a held-out subset of \textit{LightProp}, consisting of $1,000$ randomly selected, unseen objects.

\noindent\textbf{Normal.} 
Following prior works~\cite{fu2024geowizard,ye2024stablenormal}, we report the comparison results in mean and median angular errors, and accuracy within various angular thresholds. Since we observe promising accuracy within the commonly used thresholds from 5\degree to 30\degree, we further report the accuracy under a finer threshold of 3\degree. As shown in \cref{tab:comparison_full_normal}, our method outperforms baselines across all metrics, particularly under finer thresholds, clearly showing the effectiveness. 

\noindent\textbf{Materials and Relighting.} 
Following previous works, we calculate PSNR and RMSE for albedo, roughness, and metallic maps, and evaluate relit images using PSNR, SSIM, and LPIPS~\cite{zhang2018lpips}. We also report the average time per frame, calculated by measuring the total time to render $120$ relit frames from a single input image and dividing by the number of frames. As shown in \cref{tab:comparison_full_pbr_relit}, our method shows a clear improvement over baselines.
These results demonstrate the effectiveness and efficiency of our approach in predicting accurate material properties and rendering faithful relighting images.

\subsection{Qualitative Evaluation}
We present qualitative comparison results on both the unseen Objaverse subset and in-the-wild images. More visual results are given in \emph{appendix}.

\noindent\textbf{Normal.}
As shown in \cref{fig:normal_compare_main}, our method produces sharp, coherent normal maps while preserving surface details. For instance, in the cow case, our method accurately captures the normal variations around the ears. In the robot example, other methods tend to produce over-smoothed or inaccurate normal, while ours demonstrates a clear advantage in capturing complex surface geometries.
Please refer to \cref{fig:normal_compare_supp_eval} for more examples.

\noindent\textbf{PBR Materials.}
As shown in \cref{fig:pbr_compare_main}, our approach generates more accurate PBR materials than baselines. Baseline methods fail to remove highlights in their albedo maps, while our approach produces smooth base colors regardless of the illumination conditions of input images. Also, our method is more robust at distinguishing metal and nonmetal materials, while baselines are prone to reflective parts or fail to locate the metallic regions.
More examples can be found in \cref{fig:pbr_compare_supp_1,fig:pbr_compare_supp_2}.

\noindent\textbf{Image Relighting.}
As shown in \cref{fig:relit_compare_main}, our approach generates realistic lighting effects and retains details such as Chinese characters in the last example. In contrast, without underlying physical properties, DiLightNet and IC-Light tend to generate over-saturated images, while others are limited in eliminating highlights and shadows from the input image. 
Video comparisons are provided in our \emph{\href{https://projects.zxhezexin.com/neural-lightrig}{project page}}.
In the appendix, we provide more relighting comparisons in \cref{fig:relit_compare_supp} and more relighting results of our method in \cref{fig:various_relighting_supp_1,fig:various_relighting_supp_2}.

\begin{table}[t]
\centering
  \caption{Effects of condition strategies in multi-light diffusion.}
  \label{tab:ablation_mld_arch_mldstage}
  \scriptsize
  \begin{tabular}{l|ccc}
    \toprule
    \textbf{} & \textbf{PSNR~$\uparrow$} & \textbf{SSIM~$\uparrow$} & \textbf{LPIPS~$\downarrow$}
    \\
    \midrule
    Concatenation  & 19.32 & 0.8597 & 0.0909
    \\
    Reference Attention & 19.87 & 0.8691 & 0.0829
    \\
    Concatenation + Reference Attention  & \textbf{20.01} & \textbf{0.8718} & \textbf{0.0815}
    \\
  \bottomrule
\end{tabular}
\end{table}
\subsection{Ablation Study}
\label{sec:ablation}
Due to the expensive training cost of the full model, we use smaller models for the following ablation experiments.

\begin{figure}[t]
    \centering
    \includegraphics[width=0.97\linewidth]{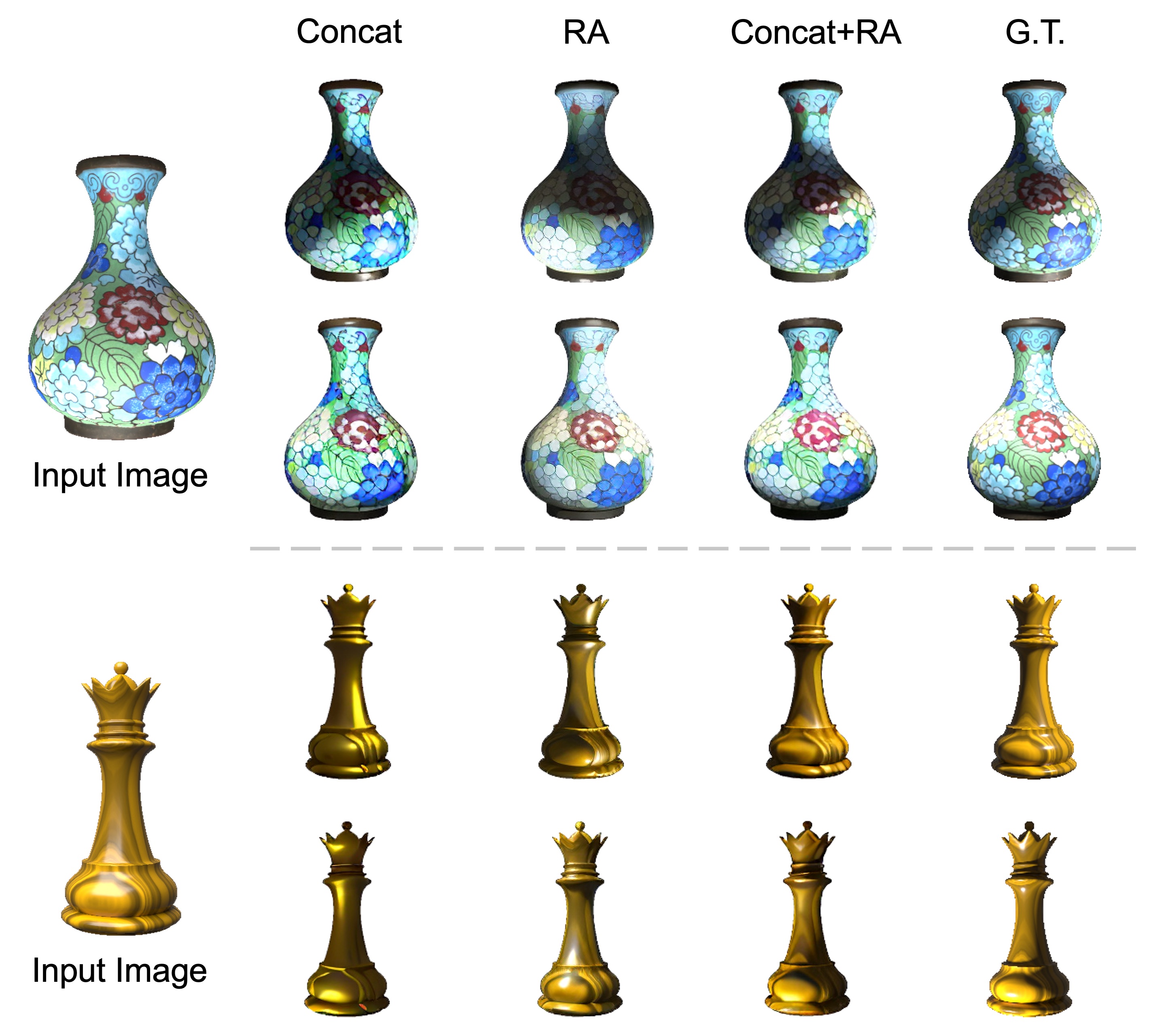}
    \caption{Visualization of different conditioning strategies in multi-light diffusion. \textit{Concat} stands for concatenation. \textit{RA} stands for reference attention.}
    \label{fig:ablation_diffusion}
    \vspace{-1mm}
\end{figure}

\begin{table*}[t]
\centering
  \caption{Effect of the number of multi-light images on the performance of the large G-buffer model. }
  \label{tab:ablation_recon_ref_reconstage}
  \scriptsize
  \begin{tabular}{c @{\hspace{8mm}}| cc | cc | cc | ccccc}
    \toprule
    \textbf{Number of} & \multicolumn{2}{c|}{\textbf{Albedo}} & \multicolumn{2}{c|}{\textbf{Roughness}} & \multicolumn{2}{c|}{\textbf{Metallic}} & \multicolumn{5}{c}{\textbf{Normal}}
    \\
    \textbf{Light Images}& \textbf{PSNR~$\uparrow$} & \textbf{RMSE~$\downarrow$} & \textbf{PSNR~$\uparrow$} & \textbf{RMSE~$\downarrow$} & \textbf{PSNR~$\uparrow$} & \textbf{RMSE~$\downarrow$} & \textbf{MAE~$\downarrow$} & \textbf{5\degree~$\uparrow$} & \textbf{7.5\degree~$\uparrow$} & \textbf{11.25\degree~$\uparrow$} & \textbf{22.5\degree~$\uparrow$}
    \\
    \midrule
    0  & 22.22 & 0.082 & 20.99 & 0.104 & 18.56 & 0.136 & 7.563 & 45.846 & 61.425 & 76.948 & 95.488
    \\
    3  & 23.72 & \underline{0.068} & 23.89 & 0.075 & \underline{20.66} & \underline{0.106} & 4.763 & 68.344 & 80.896 & 89.959 & 97.928
    \\
    6  & \underline{23.82} & \underline{0.068} & \underline{24.19} & \underline{0.072} & 20.64 & \underline{0.106} & \underline{4.275} & \underline{72.777} & \underline{83.997} & \underline{91.730} & \underline{98.312}
    \\
    9  & \textbf{23.90} & \textbf{0.067} & \textbf{24.36} & \textbf{0.069} & \textbf{20.74} & \textbf{0.105} & \textbf{4.059} & \textbf{74.720} & \textbf{85.092} & \textbf{92.330} & \textbf{98.431}
    \\
  \bottomrule
\end{tabular}
\end{table*}

\begin{table*}[t]
\centering
  \caption{Effect of augmentation strategy on the large G-buffer model.}
  \label{tab:ablation_recon_aug_full}
  \scriptsize
  \begin{tabular}{l@{\hspace{4mm}} | cc | cc | cc | ccccc}
    \toprule
   & \multicolumn{2}{c|}{\textbf{Albedo}} & \multicolumn{2}{c|}{\textbf{Roughness}} & \multicolumn{2}{c|}{\textbf{Metallic}} & \multicolumn{5}{c}{\textbf{Normal}}
    \\
    & \textbf{PSNR~$\uparrow$} & \textbf{RMSE~$\downarrow$} & \textbf{PSNR~$\uparrow$} & \textbf{RMSE~$\downarrow$} & \textbf{PSNR~$\uparrow$} & \textbf{RMSE~$\downarrow$} & \textbf{MAE~$\downarrow$} & \textbf{5\degree~$\uparrow$} & \textbf{7.5\degree~$\uparrow$} & \textbf{11.25\degree~$\uparrow$} & \textbf{22.5\degree~$\uparrow$}
    \\
    \midrule
    w/o augmentation  & 21.69 & 0.087 & 20.46 & 0.110 & 16.61 & 0.179 & 7.080 & 52.235 & 67.032 & 80.115 & 94.802
    \\
    w/ augmentation & \textbf{22.36} & \textbf{0.081} & \textbf{21.39} & \textbf{0.099} & \textbf{18.81} & \textbf{0.135} & \textbf{6.342} & \textbf{55.893} & \textbf{70.326} & \textbf{82.848} & \textbf{96.230}
    \\
  \bottomrule
\end{tabular}
\end{table*}
\noindent\textbf{Conditioning Strategy for Multi-Light Diffusion.}
We explore three different settings, concatenation, reference attention (\textit{RA}), and our hybrid approach. The quantitative analyses are given in \cref{tab:ablation_mld_arch_mldstage}. As shown in \cref{fig:ablation_diffusion}, while \textit{Concat} captures correct highlights and shadows, it often results in over-saturated colors or inaccurately rendered surface textures, as seen in the excessive brightness on the vase and inconsistent color tones on the chess piece.
\textit{RA}, on the other hand, fails to reflect faithful lighting effects. In contrast, the hybrid approach yields the best qualitative and quantitative performances.

\begin{figure}[t]
    \centering
    \includegraphics[width=\linewidth]{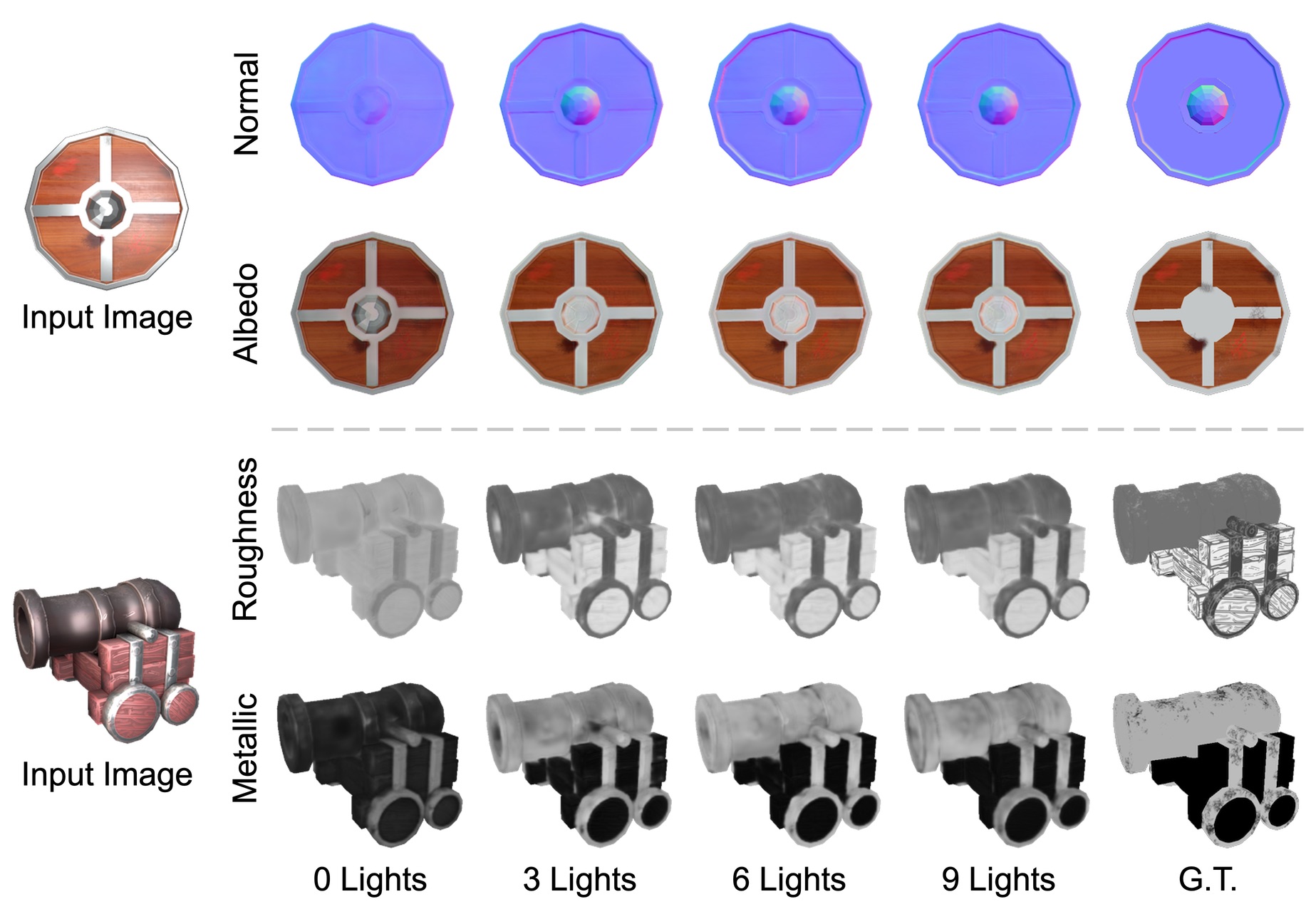}
    \caption{Visualization of using different numbers of multi-light images. We evaluate the G-Buffer prediction model with different numbers of novel-light images ($0$, $3$, $6$, and $9$) as conditions.}
    \label{fig:ablation_ref}
\end{figure}

\noindent\textbf{Number of Multi-Light Images for Prediction.}
To examine how multi-light images affect performance, we evaluate the large G-buffer model with varying numbers of rendered light images ($0$, $3$, $6$, and $9$).  As shown in \cref{tab:ablation_recon_ref_reconstage}, the performances improve sharply from $0$ to $3$ images by reducing ambiguity, and steadily improve with more provided images. The same conclusion is also observed in \cref{fig:ablation_ref}, where leveraging multi-light images yields sharper normal and better PBR maps.

\begin{figure}[t]
    \centering
    \includegraphics[width=0.9\linewidth]{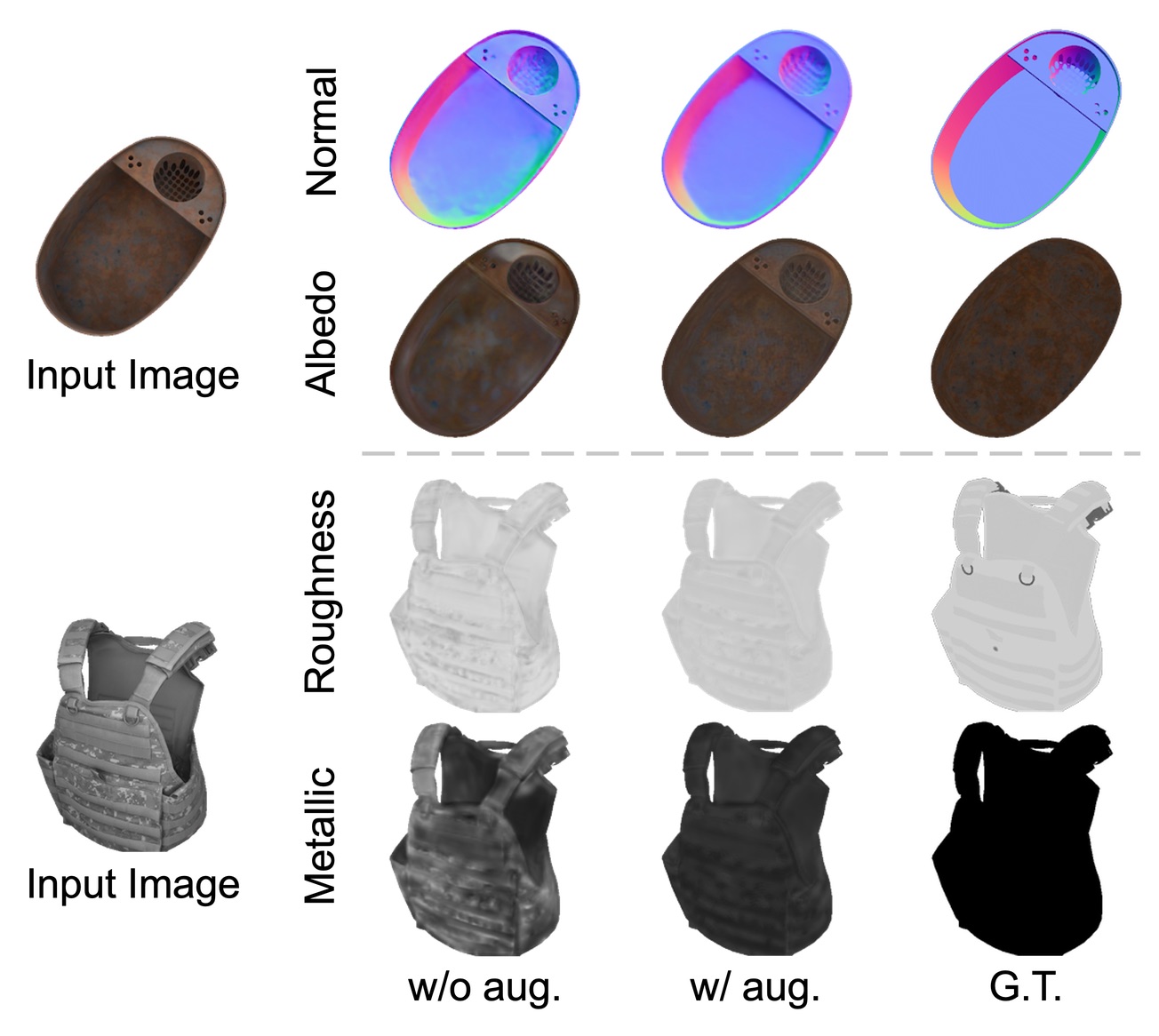}
    \caption{Visualization of the augmentation strategy.}
    \label{fig:ablation_aug}
\end{figure}

\noindent\textbf{Effects of Augmentation Strategy.}
We examine the impact of data augmentation on enhancing the robustness and generalization of the G-buffer prediction model. As shown in \cref{tab:ablation_recon_aug_full} and \cref{fig:ablation_aug}, the proposed augmentation strategy improves the model’s ability to produce consistent and accurate outputs, demonstrating increased invariance to artifacts introduced by the multi-light diffusion model. This augmentation effectively bridges the gap caused by noise, color inconsistencies, and other disturbances.

%% file: Sections/6_conclusion.tex
\section{Conclusion}
In this work, we present \emph{Neural LightRig}, a framework capable of estimating accurate surface normals and PBR materials from a single image. Leveraging a multi-light diffusion model, we generated consistent relit images under various directional light sources. These generated images significantly reduce the inherent ambiguity when estimating surface properties, serving as enriched conditions for the G-Buffer prediction model. Extensive experiments demonstrate that our method achieves significant improvements in both quality and generalizability. Future work will focus on extending this approach to more complex scenes and integrating it with 3D reconstruction systems.

%% file: Sections/7_supp.tex
\clearpage
\appendix
\section*{\centering \Large \textbf{Appendix}}

\section{Dataset Details}
In the main paper, we provided an overview of the \emph{LightProp} dataset, designed specifically to address the challenges of learning robust multi-light image generation and geometry-material estimation. Here, we detail the data curation and rendering configurations.

\subsection{Data Curation}
Objaverse~\cite{objaverse} originally contains around $800,000$ synthetic objects across various categories and styles. To ensure high-quality content for \emph{LightProp}, we implemented a rigorous curation process. First, we filtered out objects with extreme thinness or unbalanced proportions, such as objects with large surface areas but minimal thickness or depth, which often distort lighting interactions and hinder effective learning. Additionally, we excluded objects that originated from 3D scans or those representing entire scenes, as these typically contain irrelevant environmental details that are less suitable for our framework. Finally, objects lacking essential PBR material maps (albedo, roughness, and metallic maps) were removed to ensure comprehensive material data for training. This selection process resulted in a refined subset of around $80,000$ high-quality objects for \emph{LightProp}.

\subsection{Rendering Setup}
The \textit{LightProp} dataset is created using the Cycles rendering engine in Blender~\cite{blender}, with each image generated at 128 samples per pixel and accelerated using CUDA.
To introduce diversity in object orientation and perspective, each object is rendered from five distinct viewpoints: a front view, a right view, a top view, and two random views sampled on a surrounding sphere. For each viewpoint, we apply five distinct lighting conditions, comprising a point light, an area light, and three HDR environment maps randomly selected from $25$ high-quality maps.
To set up our directional lighting, we position eight lights around the camera and place one additional light directly at the camera's position. The lighting orientations are parameterized by spherical coordinates $\theta$ and $\varphi$, specifically configured as:
\begin{align}
    \theta_i &= i \cdot \frac{\pi}{4} \quad \text{for } i = 0, 1, \dots, 8, \\
    \varphi_i &= \{1, 2, 1, 2, 1, 2, 1, 2, 0\} \cdot \frac{\pi}{6}.
\end{align}
This arrangement ensures diverse lighting directions to enhance shading and reflectance variations in multi-light images, which are essential for accurate geometry and material estimation.
In addition to the multi-light images, each object view is paired with ground-truth G-buffer maps, including surface normals, albedo, roughness, and metallic maps. These G-buffers, rendered via Blender’s physically-based pipeline, provide the necessary supervision for training in surface normal and PBR material prediction.

\section{Implementation Details}
\subsection{Multi-Light Diffusion}
We build our multi-light diffusion model on top of Stable Diffusion v2-1\footnote{https://huggingface.co/stabilityai/stable-diffusion-2-1}. As discussed in the main paper, we adopt a two-phase training scheme to adapt this pre-trained model for multi-light image generation.
In the initial phase, we tune the first convolution layer, all parameters in the self-attention layers, and only the key and value parameters in the cross-attention layers. This phase runs for $80,000$ steps with a peak learning rate of $1\times10^{-4}$ and a total batch size of $128$, following a cosine annealing schedule with $2,000$ warm-up steps. We use the AdamW optimizer with $\beta_1=0.9$, $\beta_2=0.999$, and a weight decay of $0.01$, and enable bf16 mixed precision to accelerate the training. Additionally, we apply gradient clipping with a maximum norm of $1.0$ to stabilize training and incorporate classifier-free guidance, with a probability of dropping the conditioning set to $0.1$.
In the following phase, we further fine-tune the full model for another $80,000$ steps at a significantly lower peak learning rate of $5\times10^{-6}$ with the same training particulars.
Both of the two phases are trained with an input image resolution of $256 \times 256$, and a multi-light output of $768 \times 768$. In total, the complete training process of our multi-light diffusion model takes approximately $2.5$ days on $32$ NVIDIA A100 (80G) GPUs.

\subsection{Large G-Buffer Prediction Model}
\noindent\textbf{Architecture.}
Our large G-buffer prediction model takes as input a single image with $4$ channels (including alpha), combined with multi-light images comprising $9$ lighting conditions, each with $3$ channels, resulting in a total of $4 + 9 \times 3 = 31$ input channels. The output consists of $8$ channels, representing the surface normals, albedo, roughness, and metallic maps ($3$, $3$, $1$, and $1$ channel, respectively).
The regression U-Net architecture comprises four down-sampling blocks with progressively increasing channels of $224$, $448$, $672$, and $896$, followed by a bottleneck block with $896$ channels, and then four up-sampling blocks with correspondingly decreasing channels of $896$, $672$, $448$, and $224$. Each block contains two residual layers with Group Normalization (using $32$ groups), and SiLU activation. Attention mechanisms, implemented in a pre-norm style~, are applied in all but the first down-sampling block and the last up-sampling block, using an attention head dimension of $8$. Within each block, up-sampling and down-sampling are performed via a convolutional layer placed after the two residual layers.
To encode the spherical coordinates $\{\theta^i,\varphi^i\}$ associated with each lighting condition, we employ sinusoidal embeddings. Each scalar $\theta$ or $\varphi$ is projected to a higher dimension of $d_{scalar}=224$ and we concatenate these projected vectors into a single $9\times2\times224=4032$ dimensional vector, which is subsequently embedded by a $2$-layer MLP, producing an illumination embedding with a final dimensionality of $d_{emb}=896$. This embedding is modulated to each block in the U-Net with adaptive group normalization.
For the smaller models in our ablation study, we use a U-Net with down-sampling blocks at $128$, $256$, $384$, and $512$ channels, mirrored in the up-sampling blocks, along with a $512$-channel bottleneck block.

\noindent\textbf{Training Details.}
We apply weighted loss contributions to balance $\mathcal{L_{\text{normal}}}$ and $\mathcal{L_{\text{PBR}}}$. Specifically, we set a $4:1$ ratio for surface normals relative to PBR materials. Additionally, we apply a stabilization factor of $\lambda_1 = 0.25$ for the MSE term in $\mathcal{L}_{\text{normal}}$, as outlined in the main paper.
Given the computational demands of high-resolution feature maps, especially with attention layers, we employ a two-phase training strategy, gradually transitioning from low to high resolutions. In the initial phase, we train at a resolution of $256 \times 256$ to establish core feature representations, running for $60,000$ steps with a batch size of $128$. This phase includes $1,500$ warm-up steps, a peak learning rate of $1 \times 10^{-4}$, and a weight decay of $0.01$, using a cosine annealing schedule and the AdamW optimizer with $\beta_1=0.9$ and $\beta_2=0.999$. Training on $32$ NVIDIA A100 (80G) GPUs, this phase completes in approximately $20$ hours. Following this foundational phase, we move to a higher resolution of $512 \times 512$, allowing the model to capture finer details essential for precise geometry and material predictions. This fine-tuning phase involves a reduced learning rate of $2 \times 10^{-5}$ and runs for an additional $30,000$ steps on the same setup of $32$ NVIDIA A100 (80G) GPUs, completing in approximately $7$ days. All other training parameters are kept consistent with the initial phase.

\noindent\textbf{Augmentation Details.}
In the main paper, we introduced the augmentations to bridge the gap between our multi-light diffusion model and the large G-buffer prediction model.
For \textit{Random Degradation}, we down-sample each multi-light image to a lower resolution uniformly sampled from $\mathcal{U}(128,256)$ and then up-sample it back to the original resolution of $256$. Following this, we apply grid distortion with a perturbation strength sampled from $\mathcal{U}(0.15, 0.3)$ to simulate geometrical misalignments.
For \textit{Random Intensity}, we convert the multi-light images to HSV format and adjust the brightness channel using an image-level scaling factor from $\mathcal{U}(0.9,1.3)$. Additionally, we apply pixel-level noise by scaling each pixel independently with a factor sampled from $\mathcal{N}(1,0.05)$. The input image receives a separate brightness adjustment factor sampled from $\mathcal{U}(0.9,1.1)$.
For \textit{Random Orientation}, all spherical coordinates are perturbed by an angular gaussian noise in radians. ${\theta^i}$ receive a noise sampled from $\mathcal{N}(0,0.1)$ and are wrapped with modulus $2\pi$. ${\varphi^i}$ are perturbed with noise from $\mathcal{N}(0,0.02)$ and clamped within $[0, \frac{\pi}{2}]$.
The above three augmentations are triggered independently with a probability of $0.6$.
For \textit{Data Mixing}, this augmentation is applied with a probability of $0.3$. We generate multi-light images from our diffusion model with a classifier-free guidance scale of $2.0$ over $75$ inference steps.
Additionally, inspired by prior work on multi-view reconstruction~\cite{li2024instant3d}, we shuffle the order of the multi-light images during training with a probability of $0.5$ to encourage robustness in learning features across varied lighting sequences, thereby reducing dependency on any specific lighting arrangement.

\section{Limitations}
\begin{figure}[t]
    \centering
    \includegraphics[width=\linewidth]{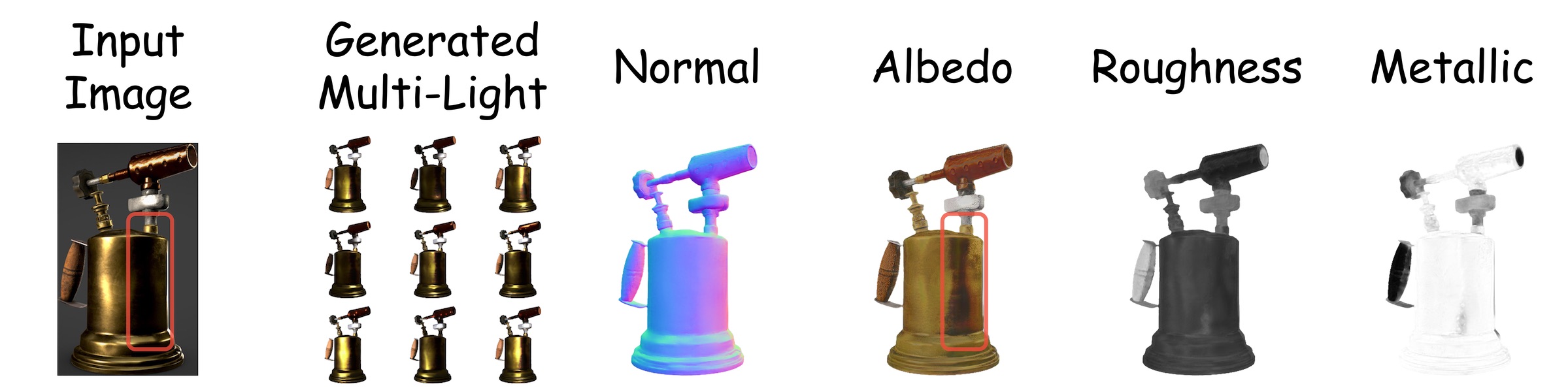}
    \caption{Failure case.}
    \label{fig:fail_supp}
\end{figure}

While our approach demonstrates strong performance, several limitations remain. First, for input images with extreme highlights or shadow areas, our method struggles to fully remove illumination effects in the predicted albedo maps, as shown in \cref{fig:fail_supp}. Additionally, the resolution of the backbone multi-light diffusion model ($256 \times 256$) limits the level of detail achievable in the generated multi-light images, subsequently constraining the final normal and material predictions. Increasing the model’s resolution could enhance the quality of the predicted surface properties. Finally, our method is currently designed for objects rather than full scenes, limiting its applicability in complex, multi-object environments.

\section{Additional Results}

\subsection{Our Results}
\begin{figure*}[h]
    \centering
    \includegraphics[width=\linewidth]{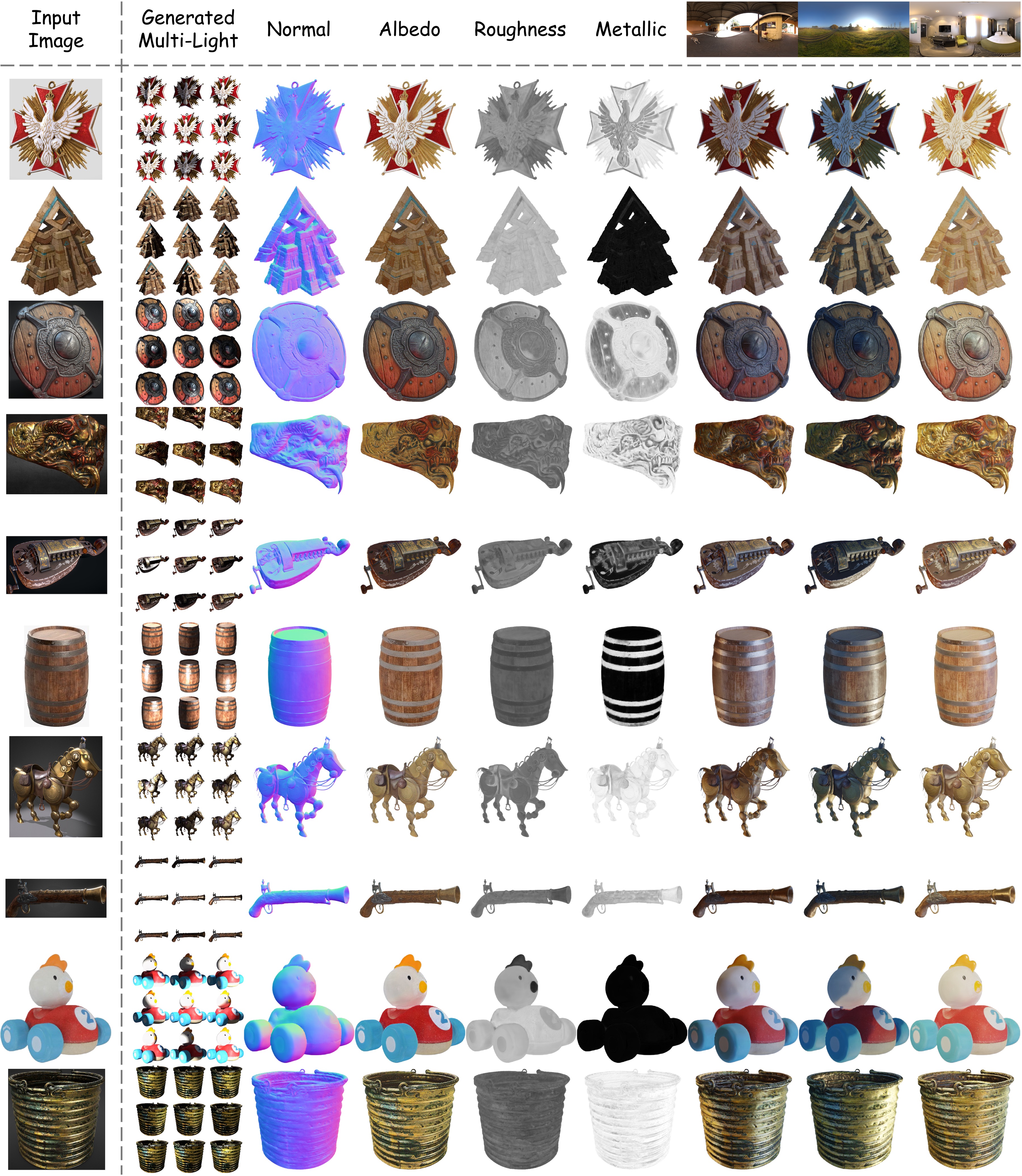}
    \caption{More results of our method.}
    \label{fig:results_supp_1}
\end{figure*}

\begin{figure*}[h]
    \centering
    \includegraphics[width=\linewidth]{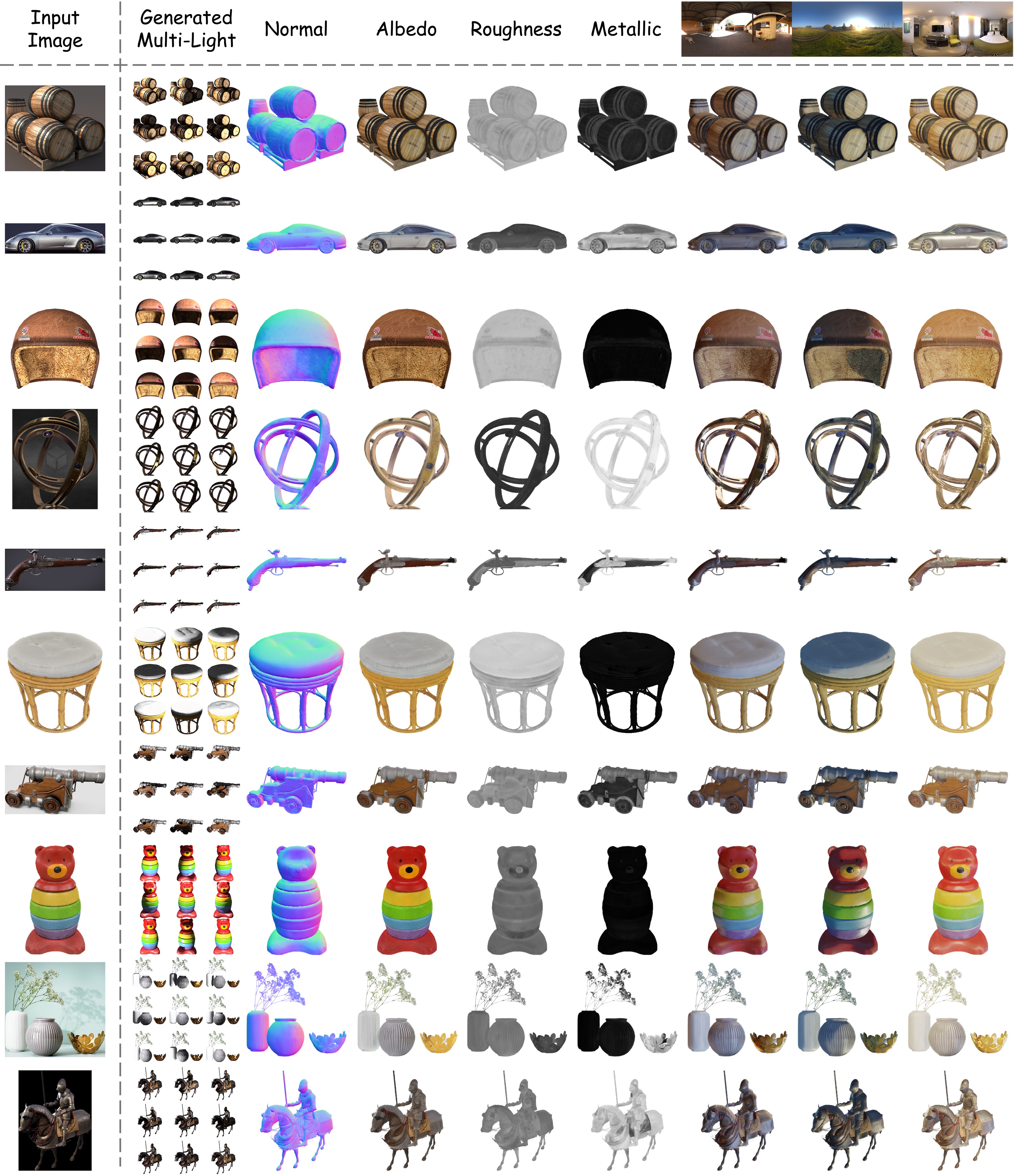}
    \caption{More results of our method.}
    \label{fig:results_supp_2}
\end{figure*}

\begin{figure*}[h]
    \centering
    \includegraphics[width=\linewidth]{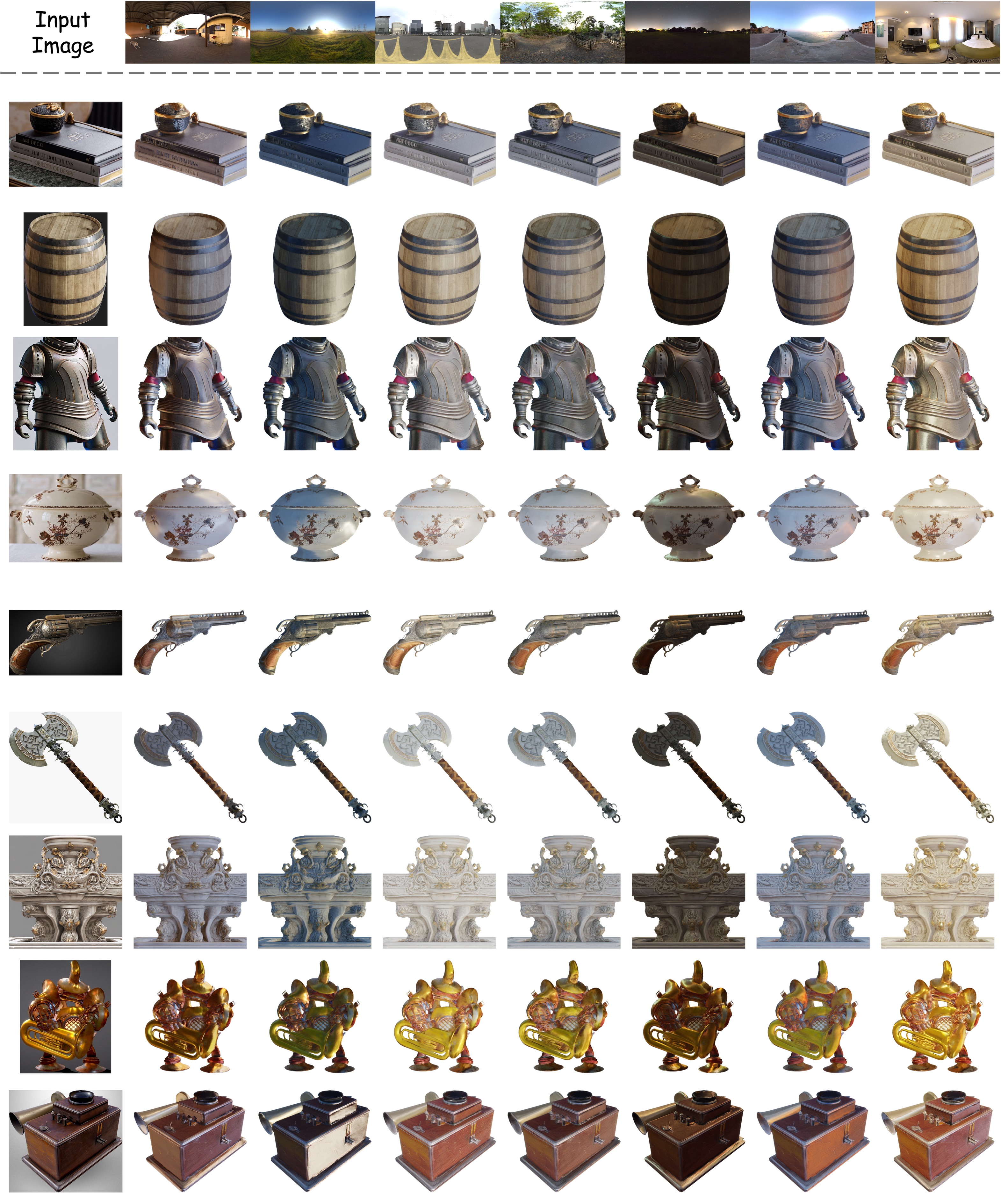}
    \caption{More single-image relighting results of our method.}
    \label{fig:various_relighting_supp_1}
\end{figure*}

\begin{figure*}[h]
    \centering
    \includegraphics[width=\linewidth]{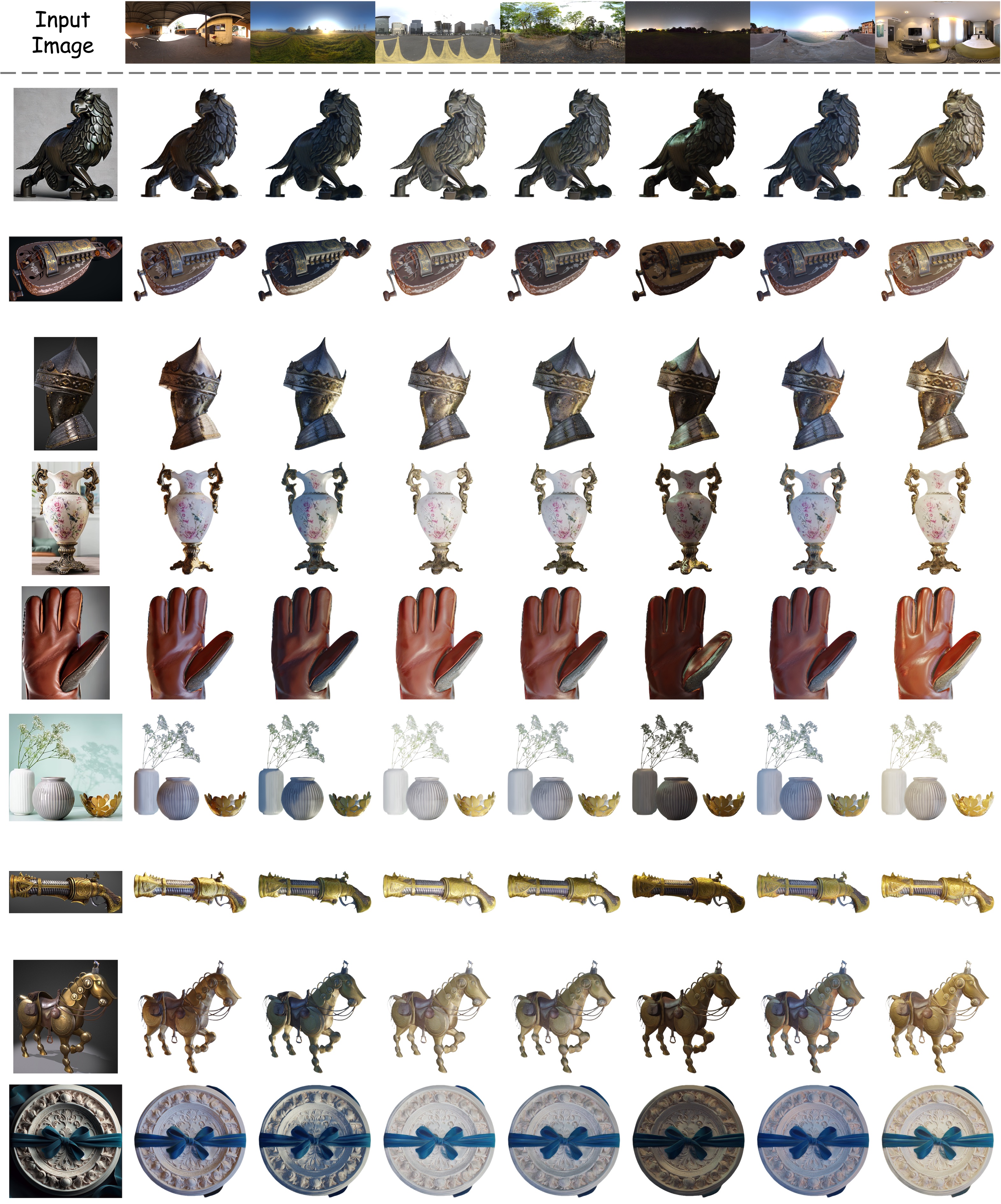}
    \caption{More single-image relighting results of our method.}
    \label{fig:various_relighting_supp_2}
\end{figure*}

\cref{fig:results_supp_1,fig:results_supp_2} present examples of our full pipeline output, including input images, generated multi-light images, estimated surface normals, PBR materials, and relit images under various environment maps. These results showcase the robustness of our approach in generating consistent geometry and material estimates and realistic relighting effects across different lighting conditions. Additionally, \cref{fig:various_relighting_supp_1,fig:various_relighting_supp_2} showcase extended single-image relighting results of our method under an even broader range of environment maps, further highlighting the model’s ability to generate high-quality, adaptable relit images across diverse lighting setups.
These results illustrate the robustness in managing various lighting conditions and further demonstrate the efficacy of our approach.

\subsection{Comparison Results}
\begin{figure*}[h]
    \centering
    \includegraphics[width=0.95\linewidth]{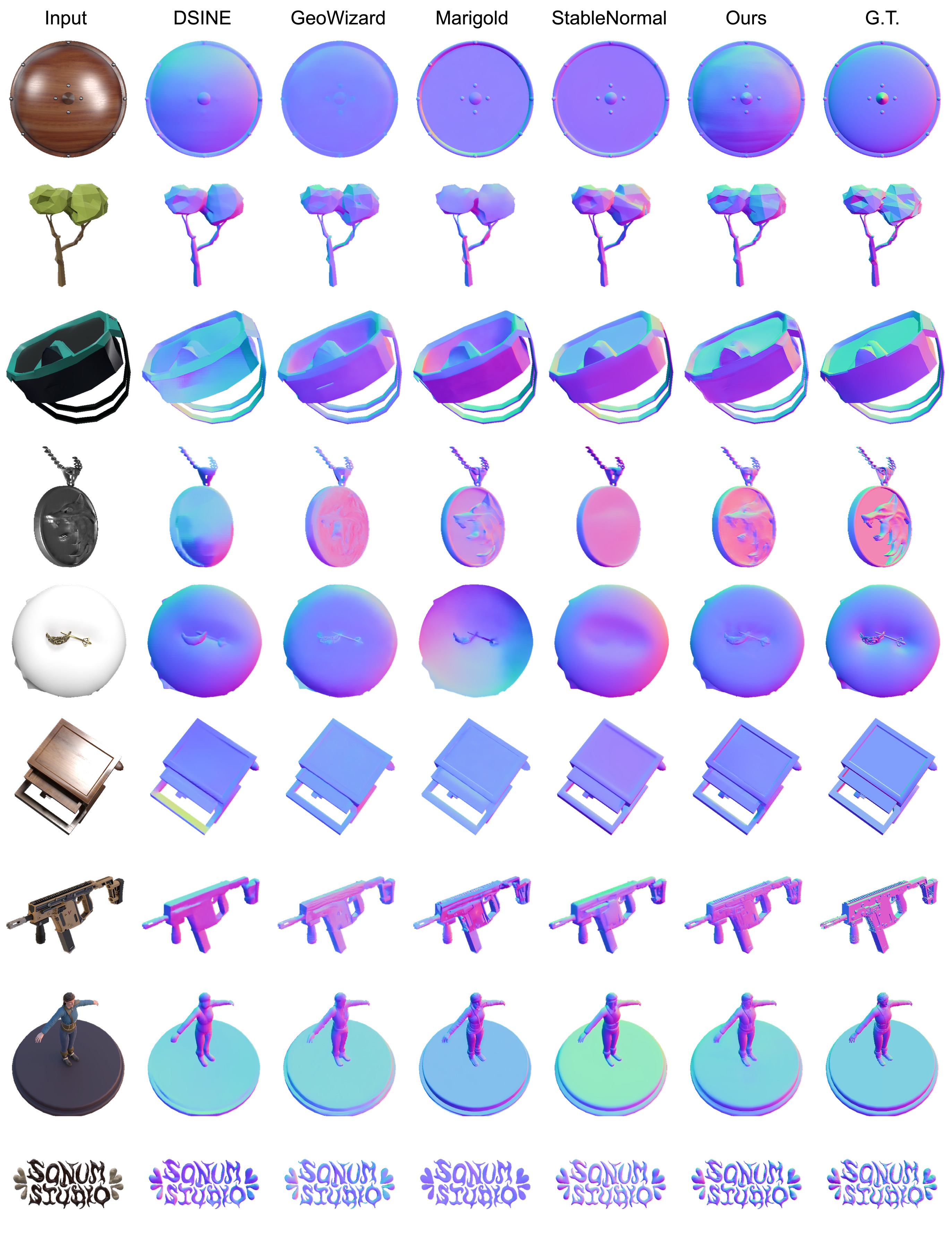}
    \caption{More comparisons on surface normal estimation.}
    \label{fig:normal_compare_supp_eval}
\end{figure*}

\begin{figure*}[h]
    \centering
    \includegraphics[width=0.9\linewidth]{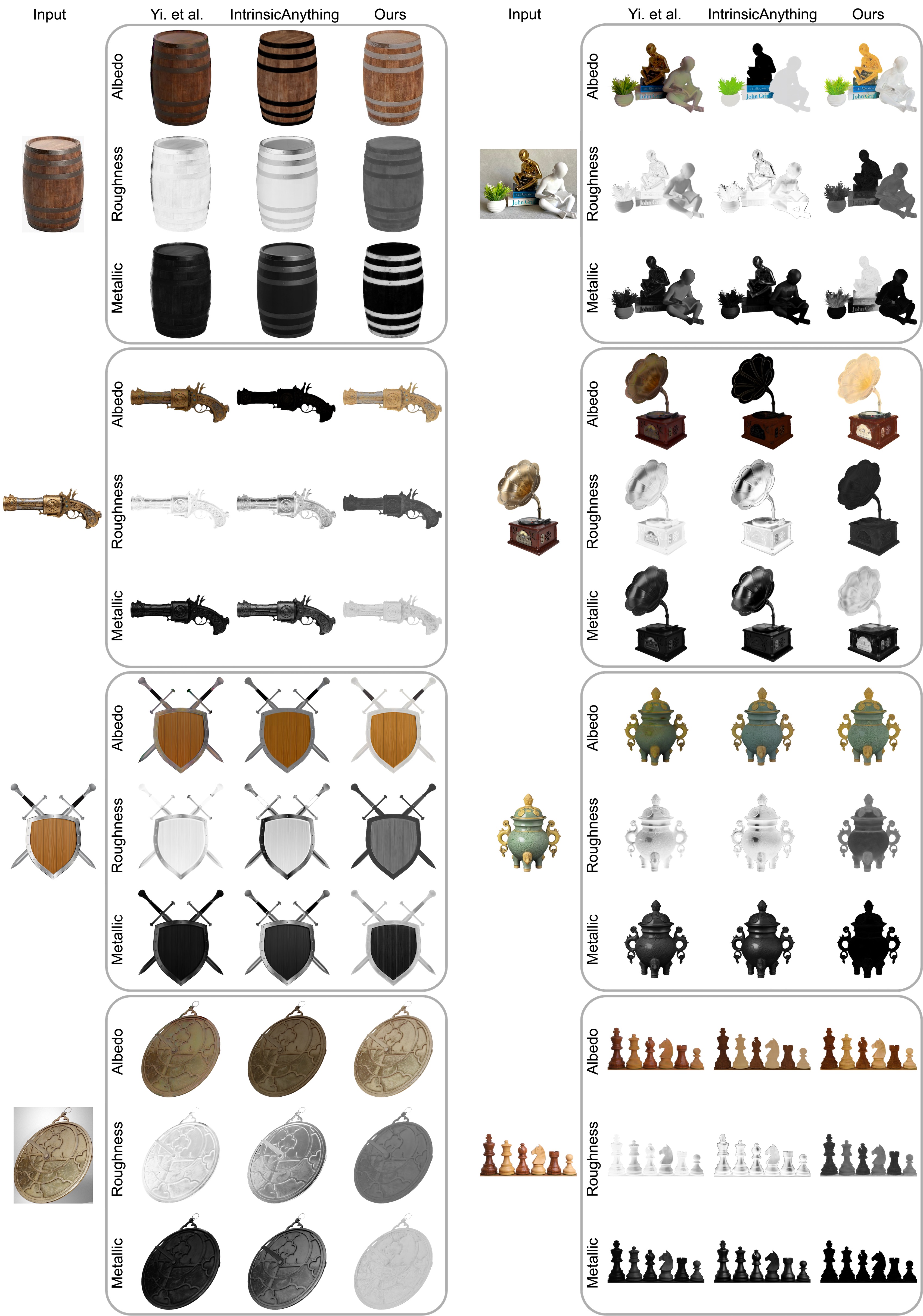}
    \caption{More comparisons on PBR material estimation.}
    \label{fig:pbr_compare_supp_1}
\end{figure*}

\begin{figure*}[h]
    \centering
    \includegraphics[width=0.9\linewidth]{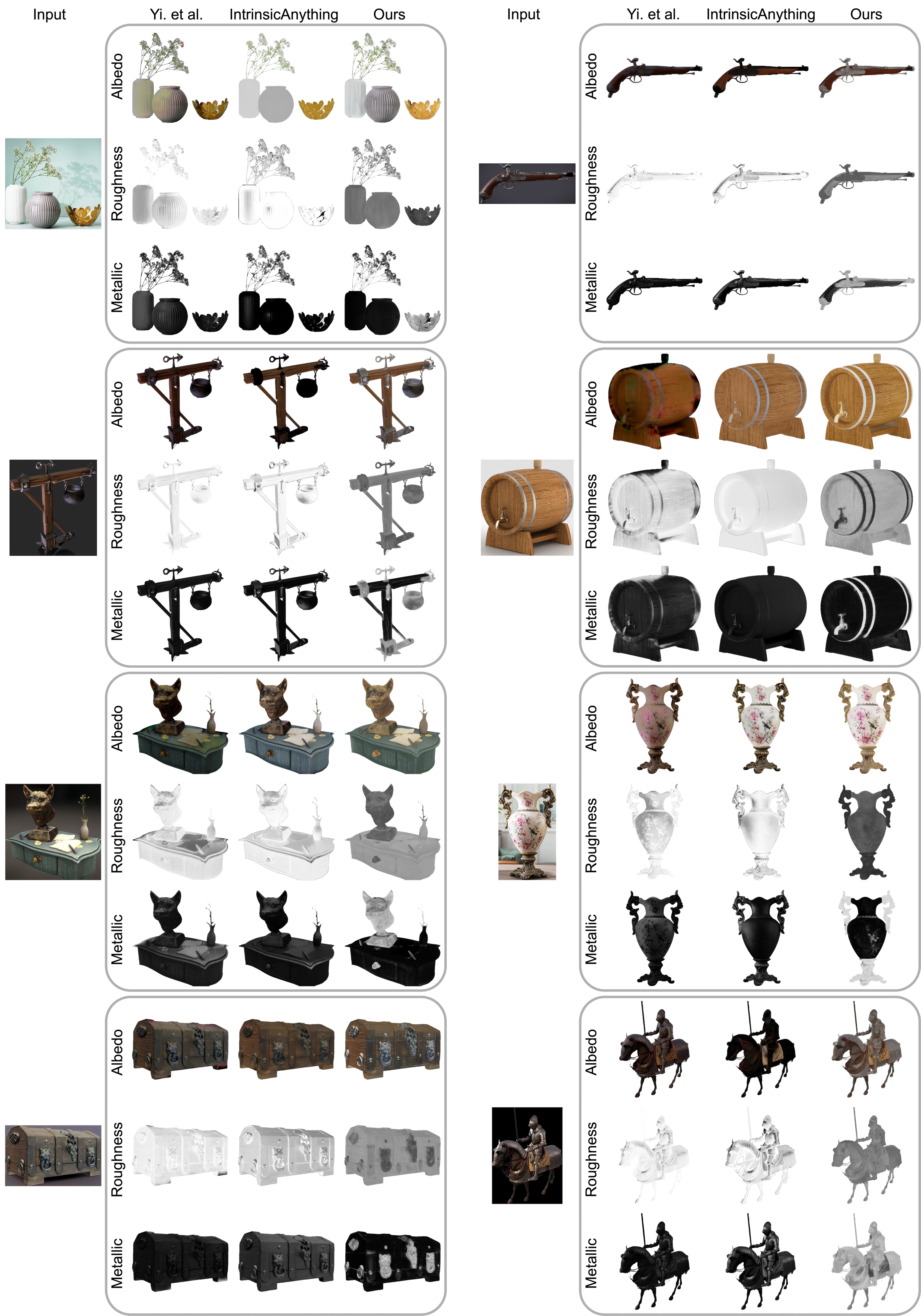}
    \caption{More comparisons on PBR material estimation.}
    \label{fig:pbr_compare_supp_2}
\end{figure*}

\begin{figure*}[h]
    \centering
    \includegraphics[width=\linewidth]{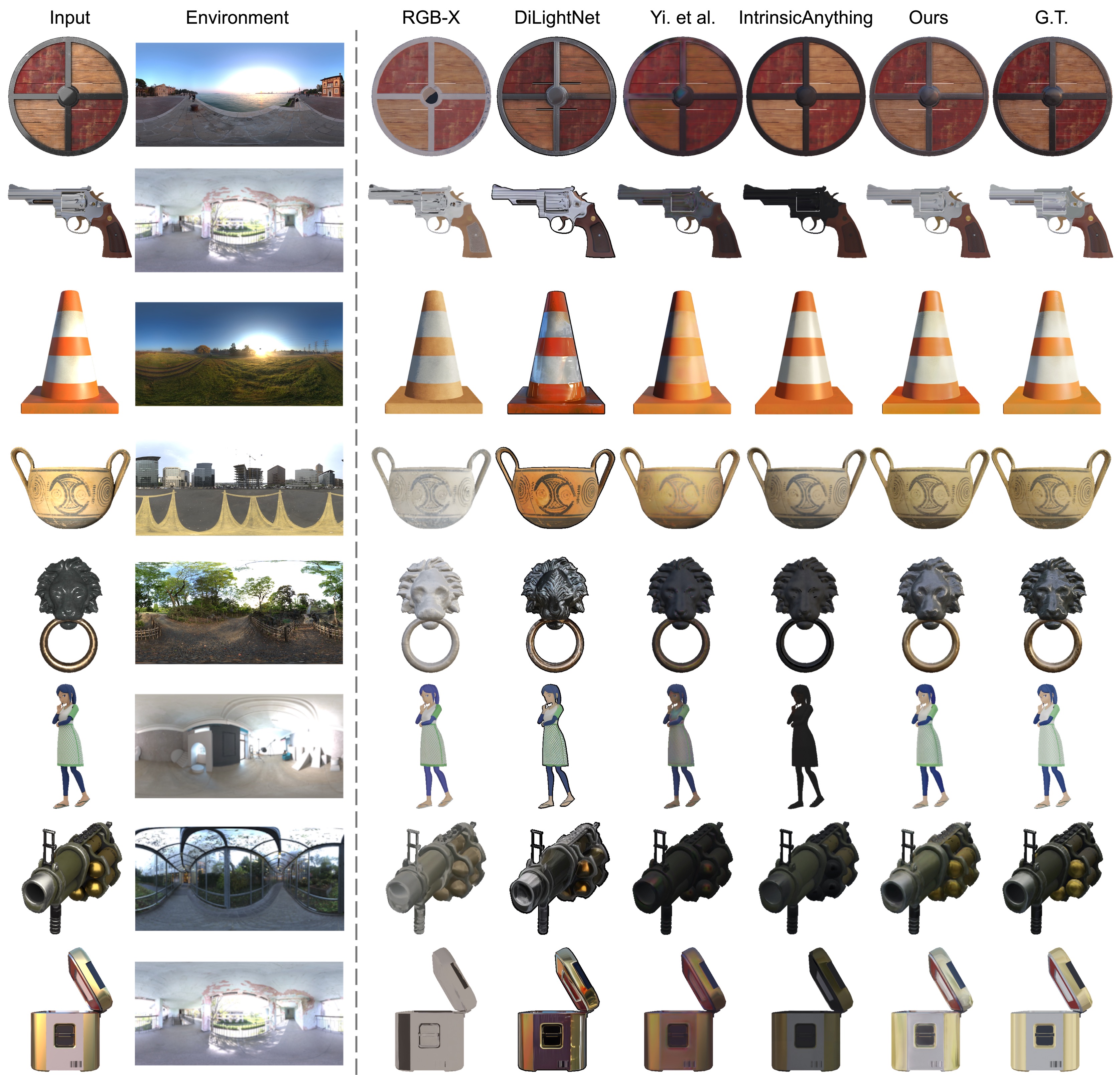}
    \caption{More comparisons on single-image relighting.}
    \label{fig:relit_compare_supp}
\end{figure*}

In \cref{fig:normal_compare_supp_eval}, \cref{fig:pbr_compare_supp_1}, \cref{fig:pbr_compare_supp_2}, and \cref{fig:relit_compare_supp} we offer more comparison results for surface normal estimation, PBR material estimation, and single-image relighting. These comparisons further demonstrate the advantages of our method over baseline approaches in accurately capturing surface details, material properties, and producing realistic relit images under diverse lighting conditions.

